\ifdefined\XeTeXversion\else\ifdefined\pdfoutput\pdfoutput=1\fi\fi
\documentclass{article} %
\usepackage{iclr2027_conference,times}

\usepackage[hyphens]{url}
\usepackage{hyperref}
\usepackage{graphicx}

\usepackage{amsmath}
\usepackage{amssymb}
\usepackage{booktabs}
\usepackage{float}   %
\usepackage{multirow}
\usepackage{caption}
\usepackage{subcaption}
\usepackage{nicefrac}
\usepackage{enumitem}
\usepackage{xcolor}
\usepackage{pgfplots}
\pgfplotsset{compat=1.18}

\definecolor{newblue}{RGB}{0,0,0}       %
\DeclareRobustCommand{\new}[1]{\textcolor{newblue}{#1}}
\definecolor{editblack}{RGB}{0,0,0}        %
\DeclareRobustCommand{\edit}[1]{\textcolor{editblack}{#1}}
\definecolor{latpurple}{RGB}{0,0,0}     %

\definecolor{paraorange}{RGB}{0,0,0}    %
\DeclareRobustCommand{\pedit}[1]{\textcolor{paraorange}{#1}}
\definecolor{routerviolet}{RGB}{0,0,0}  %
\DeclareRobustCommand{\routeredit}[1]{\textcolor{routerviolet}{#1}}
\definecolor{clrevblue}{RGB}{0,0,0}      %
\DeclareRobustCommand{\clrev}[1]{\textcolor{clrevblue}{#1}}
\definecolor{rtrrev}{RGB}{0,0,0}          %
\DeclareRobustCommand{\rrev}[1]{\textcolor{rtrrev}{#1}}
\definecolor{pfitgreen}{RGB}{0,0,0}        %
\DeclareRobustCommand{\pfit}[1]{\textcolor{pfitgreen}{#1}}
\newcommand{\syskey}[1]{\texttt{\bfseries #1}}   %
\newcommand{\sUp}{\syskey{Up}}          %
\newcommand{\sSone}{\syskey{Ref}}       %
\newcommand{\sB}{\syskey{S1}}           %
\newcommand{\sSthree}{\syskey{S3}}      %
\newcommand{\sAone}{\syskey{A1}}        %
\newcommand{\sAthree}{\syskey{A3}}      %
\newcommand{\sR}{\syskey{R}}            %
\definecolor{yongblue}{RGB}{0,0,0}         %

\newcommand{\name}{SpecRelay}
\newcommand{\SpecCloud}{\textsc{Spec-Cloud}}
\newcommand{\EdgeDirect}{\textsc{Edge-Direct}}
\newcommand{\EdgeCoD}{\textsc{Edge-CoD}}
\newcommand{\EdgeThink}{\textsc{Edge-Think}}

\title{SpecRelay: Amortize or Avoid the Round Trip in \mbox{Edge--Cloud} Speculative Decoding}

\author{Jai Jain$^{1}$\thanks{Equal contribution.}\quad Chihyo Ahn$^{1}$\footnotemark[1]\quad Yong Liu$^{1}$\quad Charles Chen$^{1}$\quad Duc N.M Hoang$^{2}$ \\
\bfseries Ajay Kumar Jaiswal$^{2}$\quad Fatih Ilhan$^{2}$\quad Han-Byul Kim$^{2}$\quad Lauren Hannah$^{2}$ \\
\bfseries Richa Dixit$^{2}$\quad Minsik Cho$^{2}$\quad Yingyan (Celine) Lin$^{1}$\thanks{Corresponding authors.}\quad Hyesoon Kim$^{1}$\footnotemark[2] \\[3pt]
{\mdseries $^{1}$Georgia Institute of Technology \quad $^{2}$Apple}
}

\iclrfinalcopy %

\begin{document}

\maketitle
\lhead{Preprint. Under review.} %

\begin{abstract}
Edge--cloud speculative decoding moves autoregressive drafting to an
edge device while a larger cloud model verifies draft tokens in
parallel\clrev{, so that edge drafting and cloud verification overlap under a
synchronous draft--verify schedule}.
Beyond low-latency local networks, however, \clrev{the next verification request of that schedule
can be sent only after the previous reply arrives, so every round still waits out a full
round trip, and throughput falls as the network RTT grows.}
\clrev{To recover this loss,} we present SpecRelay, an edge--cloud LLM
inference system that addresses this cost at two levels: it amortizes
round trips when cloud verification is necessary and avoids them when
local reasoning suffices. For cloud-bound queries, an \clrev{asynchronous}
verification pipeline keeps fixed-budget draft trees in flight \clrev{while
earlier replies are still outstanding, and the cloud} rejects stale requests
before the target-model forward pass (\clrev{the \textit{asynchronous schedule}}).
At the query level, an RTT-conditioned router (\clrev{the \textit{Latency-Aware
Router}}) estimates quality and completion latency for three on-device modes
(direct generation, Chain-of-Draft, and full reasoning) and cloud speculative
decoding, selecting among them under a configurable accuracy--latency
objective \clrev{without adding latency during inference, since the router runs
concurrently with the draft-model prefill}.
\clrev{The \textit{asynchronous schedule}} alone improves decoding throughput
by up to \clrev{50\%} over a reproduced synchronous baseline \clrev{across multiple benchmarks}.
\clrev{At matched target-model accuracy and 250~ms of added round-trip latency, the full system with the router delivers
\(1.95\times\) and \(1.55\times\) the throughput of the same baseline on
SPROUT and OlympiadBench, respectively, while sending only 47\% and 39\% of the queries to the cloud.}
\end{abstract}
\section{Introduction}
\label{sec:introduction}

Edge--cloud speculative decoding splits generation between a small draft
model on the edge device and a larger target model on the cloud server
\cite{leviathan2023fast,chen2023accelerating}. In each verification
round, the edge proposes candidate tokens organized as a sequence or,
more generally, a tree. The cloud server checks these candidates in
parallel and returns the longest accepted path; unmatched speculative
work is discarded. \clrev{Draft length (the depth of the draft tree sent per round trip)} therefore creates a tradeoff between
round-trip frequency and wasted speculative computation. Short \clrev{drafts}
limit the amount of work discarded after a mismatch, but they require
frequent edge--cloud round trips. Longer \clrev{drafts} can amortize a round
trip over more tokens if they are accepted, but an early mismatch can
discard a larger portion of the speculative work.

\clrev{In an edge--cloud setup,} the impact of this tradeoff depends on network latency. SpecEdge
demonstrated this design in a low-latency setting, with a round-trip time
(RTT) of roughly 14~ms \cite{park2025specedge}. In contrast, cellular,
metropolitan Wi-Fi, and \pfit{low-Earth-orbit satellite} paths commonly
incur RTTs of tens of milliseconds\pfit{, with} tail latencies
\pfit{of} hundreds of milliseconds
\cite{9322406,10.1145/3387514.3405882,
10.1145/3452296.3472923,michel2022starlink,mohan2024starlink}.
In a synchronous schedule, the edge must wait for the cloud reply before
it can safely advance, so every verification round adds another RTT.
Slow and variable links therefore raise two questions: \clrev{how to hide a longer RTT when the cloud is used}, and when to avoid the
\pfit{cloud altogether}.

We present \name{}, an edge--cloud LLM inference system that
addresses both questions: \emph{amortize} the round trip when cloud
verification is necessary and \emph{avoid} it when local reasoning
suffices. For cloud-bound queries, an \clrev{anchor-hedged asynchronous} verification
pipeline keeps multiple fixed-budget draft trees in flight. 
\clrev{When the edge sends a tree, it tags the tree with the committed prefix it was drafted from,
and the server verifies a tree only if that prefix is still current, discarding stale
trees before any target-model execution. On each reply, the edge adopts
the server's verdict and updates its draft trees to match: draft tokens that no longer
align with the accepted path are demoted, and drafting continues from those that do
(\emph{lineage-preserving rebuild}).} This overlaps
successive verification rounds without modifying either language model.
\clrev{Between successive trees, the edge additionally hedges across the $K$ most likely
continuation anchors and grows a $K$-times wider tree from them, which raises the
probability that the next tree is accepted; because the server forwards only the
subtree under the correct anchor, its verification cost is unchanged. We evaluate this
continuation hedging under both the synchronous and the asynchronous schedule.}

At query granularity, an RTT-conditioned router selects among on-device
direct generation, Chain-of-Draft \cite{xu2025chain}, full reasoning,
and cloud speculative decoding.
A frozen query encoder and nonparametric
estimators predict per-mode quality and output length; measured hardware
profiles and the current RTT estimate convert these predictions into
completion-latency estimates. Because RTT affects only the cloud path,
the same query can use remote verification on a fast link and additional
local test-time compute on a slow one. One fitted router covers the
whole RTT range without retraining. This extends content-based routing
\cite{ong2025routellm,chen2024frugalgpt,ding2024hybrid,
somerstep2025carrot} with runtime network state
and a multi-level local reasoning action space, while leaving the
language models unchanged.

In summary, our contributions are as follows:
\begin{itemize}[leftmargin=1.2em,labelsep=0.5em,itemsep=0pt,topsep=0pt]  %
    \item \textbf{Asynchronous cloud verification.}
We introduce \clrev{an asynchronous, prefix-gated} protocol that overlaps successive
verification rounds and prevents stale speculative requests from
consuming a target-model forward pass.
\clrev{It needs no RTT-specific tuning---one configuration serves every network condition---and yields higher throughput.}

\item \textbf{\clrev{Continuation hedging with lineage-preserving rebuild.}}
\clrev{At each boundary between successive draft trees, the edge hedges across the $K$ most
likely continuation anchors at no extra verification cost, and lineage-preserving rebuild
continues drafting from \pfit{the surviving lineage}.}

\item \textbf{RTT-conditioned test-time-compute routing.}
\clrev{A lightweight per-query router that selects among three on-device reasoning
budgets and cloud speculative decoding by predicted per-mode quality and latency at the current RTT estimate:
the lightest mode predicted to answer correctly.
Its $0.6$B encoder runs \rrev{concurrently with} the draft model's prefill and adds \rrev{near-zero (${<}4$~ms)} latency.}

\end{itemize}

\section{Preliminaries and Related Work}\label{sec:prelim}
\paragraph{Speculative decoding.} Speculative decoding~\cite{leviathan2023fast,chen2023accelerating} accelerates auto-regressive generation by using a fast draft model to propose multiple tokens that a slower target model verifies in parallel. Subsequent work has improved draft quality via self-speculation~\cite{3692070.3692273,zhang-etal-2024-draft} and tree-structured drafts~\cite{miao2023specinfer,3692070.3693232}. These advances assume colocated draft and target models, so the network is not a first-order concern.

\clrev{In edge--cloud speculative decoding, a draft model $M_d$ runs on the edge and a
target model $M_t$ runs in the cloud~\cite{park2025specedge}: each round, the edge drafts a tree and
\pfit{sends it as a \emph{verification request}}; $M_t$ verifies it in one forward pass and
\emph{replies} with the longest accepted path and one bonus token. \pfit{We call this mode}
\pfit{\SpecCloud{}.} The approximate latency per accepted token is}
\begin{equation}
\color{clrevblue}
\frac{(k \cdot t_d + t_v + \tau)\,o}{\alpha\,k},
\label{eq:sync_latency}
\end{equation}
\clrev{where $k$ is the draft length, $t_d$ the per-token edge draft time, $t_v$ the cloud verification time,
$\tau$ the network round trip, $\alpha$ the per-round acceptance rate (so $\alpha k$ tokens are accepted per round), and
$o \in [\tfrac13,1]$ the degree to which edge drafting, the round trip, and cloud verification overlap
($1$ strictly in sequence, $\tfrac13$ fully overlapped).}
At small $\tau$, the cost is compute-bound; as $\tau$
grows, the schedule becomes network-bound.

\begin{figure}[t]
  \centering
  \begin{minipage}[t]{0.485\linewidth}\centering
    \includegraphics[width=0.9\linewidth]{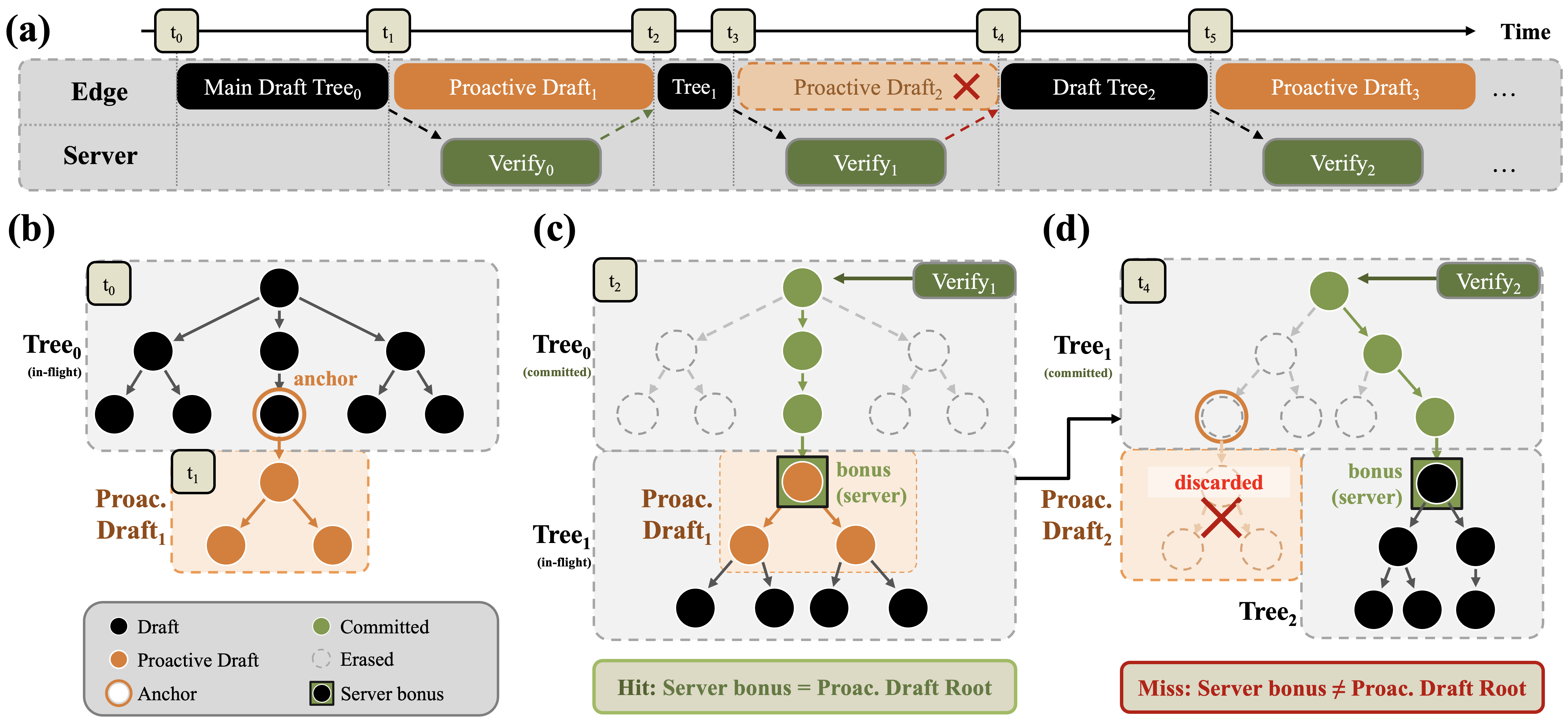}  %
    \caption{\new{Baseline SpecEdge~\cite{park2025specedge} draft tree.
  \textbf{(a)} edge and server timelines;
  \textbf{(b)} draft tree (edge) $\mathrm{Tree}_0$;
  \textbf{(c-d)} Hit(c)/Miss(d): tree after the target ($M_t$) reply.}}
  
  \vspace{-0.5em}
    \label{fig:sync_overview}
  \end{minipage}\hfill
  \begin{minipage}[t]{0.485\linewidth}\centering
    \includegraphics[width=0.9\linewidth]{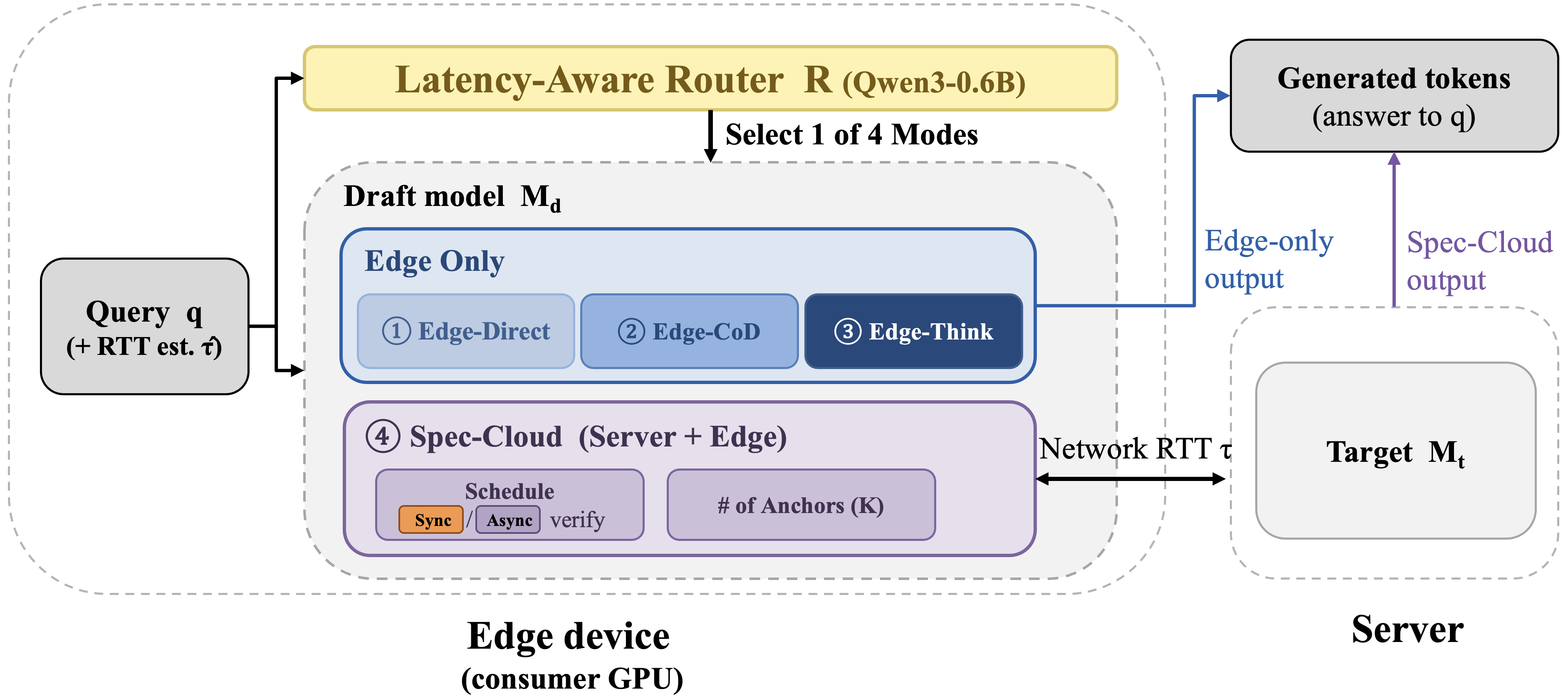}  %
    \caption{\name{} overview.}
  \vspace{-0.5em}
    \label{fig:overview}
  \end{minipage}
  \vspace{-0.9em}  %
\end{figure}

\paragraph{\clrev{Speculative decoding systems.}}
\clrev{Existing systems act on individual terms of Eq.~\eqref{eq:sync_latency}. Tree-structured
drafts~\cite{miao2023specinfer,3692070.3693232} raise $\alpha$ by verifying many continuations in one pass.
Other systems relax the lockstep draft--verify schedule in different settings: PipeSpec pipelines a hierarchy of
progressively larger models with cascading rollback over a single draft chain~\cite{mcdanel2025pipespec}; FlowSpec prunes and
expands a draft tree across a distributed model pipeline~\cite{liu2025flowspec}; and PicoSpec overlaps edge drafting,
communication, and cloud verification in a training-free edge--cloud design that drafts a short chain (four tokens) per
round~\cite{picospec2026}. Of these, only FlowSpec drafts a tree, and none targets long round trips with tree drafts
(Appendix~\ref{app:related} surveys further runahead and edge--cloud systems).
SpecEdge~\cite{park2025specedge}, our baseline (Figure~\ref{fig:sync_overview}), drafts such a tree on the edge and lowers $o$
with \emph{proactive drafting}: while the cloud verifies the current tree ($\mathrm{Tree}_0$ in Figure~\ref{fig:sync_overview}b,
sent at $t_1$), the edge drafts the next one below its most likely leaf---the \emph{anchor} from which the next tree continues---from a
single guessed bonus token; if the guess matches the
reply (Figure~\ref{fig:sync_overview}c, $t_2$), the proactive subtree is reused, and otherwise (Figure~\ref{fig:sync_overview}d,
$t_4$) it is discarded and rebuilt from the verified prefix. This can hide part of the draft time $k\,t_d$ under the wait for the reply, but
it rests on a single guess, and each round still waits $t_v+\tau$ in full (Figure~\ref{fig:async_overview}a, top): at large
$\tau$, the drafting time discarded on a miss grows with the RTT, and $o$ cannot fall below $\tfrac12$. These limits are what
\name{} addresses (\S\ref{sec:overview}--\S\ref{sec:backend}).}

\paragraph{Routing and on-device reasoning modes.}
\rrev{LLM routers decide, per query, which model or execution mode serves it, trading predicted accuracy against
cost~\cite{ong2025routellm,chen2024frugalgpt,ding2024hybrid,somerstep2025carrot}: RouteLLM learns the choice between a
strong and a weak model from preference data~\cite{ong2025routellm}; FrugalGPT cascades through models until an answer
is accepted~\cite{chen2024frugalgpt}; and CARROT estimates each model's accuracy and output length with nonparametric
regressors~\cite{somerstep2025carrot}. Subsequent work widens the choice to test-time compute: BEST-Route selects a model
together with the number of sampled responses~\cite{dingbest}, Route-to-Reason jointly selects a model and a
reasoning strategy~\cite{10.1145/3774904.3792556}, and Thinkless trains a reasoning model to decide when to think at
length~\cite{fang2025thinkless}. Device--cloud systems split the same decision across the network: Hybrid LLM routes
queries between a small and a large model~\cite{ding2024hybrid}, and Minions has an on-device model collaborate with a
cloud model on one task~\cite{narayan2025costefficient}. Of these, only device--edge routers include a communication
cost~\cite{11147210}, and none weighs the round trip $\tau$ of Eq.~\eqref{eq:sync_latency} against on-device
reasoning budgets.} \name{} conditions on both the query and \clrev{the current RTT estimate}, sending the same query to
the cloud on a fast link but serving it on device when round-trip delay is high. Its
on-device modes span reasoning budgets from direct generation to
Chain-of-Draft~\cite{xu2025chain} and full
reasoning~\cite{wei2022chain,snell2025scaling}\rrev{ (\S\ref{sec:router})}.

\section{System Overview}
\label{sec:overview}

Figure~\ref{fig:overview} shows the overall \name{} framework. 
The edge hosts a draft model \(M_d\) and router \(R\), while the cloud
hosts a larger target model \(M_t\). Given query \(q\) and the current
RTT estimate \(\hat{\tau}\), \(R\) selects one of four modes:
\EdgeDirect{}, \EdgeCoD{}, \EdgeThink{}, or
\SpecCloud{}. The first three complete generation locally, trading more tokens for
higher accuracy: \EdgeDirect{} is cheapest; \EdgeCoD{} is a
concise middle ground; and \EdgeThink{} uses full reasoning for higher
accuracy on hard queries \pfit{at the cost of} verbose decoding.
In \SpecCloud{}, \(M_d\) \pfit{drafts}
\pfit{fixed-budget trees} on the edge and \(M_t\) verifies them in the
cloud. \pfit{Neither model} is modified.

\clrev{Compared with SpecEdge, \SpecCloud{} exposes two knobs (Figure~\ref{fig:overview}): the verification schedule and
the number of anchors $K$. The schedule is synchronous, as in SpecEdge, or asynchronous, in which case the edge does not wait
for the server's reply before drafting and sending the next tree: successive fixed-budget trees are issued back to back, so
several verification requests are in flight at once (\S\ref{sec:pipeline}). At each boundary between trees, the next tree is
grown from $K$ \emph{anchors}---the $K$ most likely
continuations of the tree still being verified---rather than from one guess, so that a subsequent tree is already in place for
whichever continuation the reply confirms, and drafting continues from that lineage (\emph{lineage-preserving rebuild}, \S\ref{sec:kroot}). \name{}'s full system (\sR{} in Table~\ref{tab:setup}) runs on the
asynchronous schedule with $K{=}3$; \S\ref{sec:systems_results} evaluates the other combinations as ablations.}

\section{\new{Cloud-Path Scheduling}}
\label{sec:backend}

\clrev{This section presents the two mechanisms behind the \SpecCloud{} knobs introduced in \S\ref{sec:overview}, which address the limits of
proactive drafting noted in \S\ref{sec:prelim}: the \emph{asynchronous verification pipeline}, which overlaps the verification and round
trips of successive requests so that both $t_v$ and $\tau$ of a request are hidden behind the others in flight (lower $o$ in
Eq.~\eqref{eq:sync_latency}; \S\ref{sec:pipeline}), and
\emph{continuation hedging}, which replaces the single guess at each boundary between trees by $K$ anchors (higher $\alpha$ at the boundary;
\S\ref{sec:kroot}). Every combination of the two keeps the target model's verification rule and a fixed server-side tree budget.
Together, the two mechanisms enable \emph{lineage-preserving rebuild}:
with several hedged trees in flight, a verdict is matched against them and drafting continues from the lineage it confirms
instead of restarting from the verified prefix (\S\ref{sec:kroot}).}

\begin{figure}[t]
  \centering
  \includegraphics[width=\linewidth]{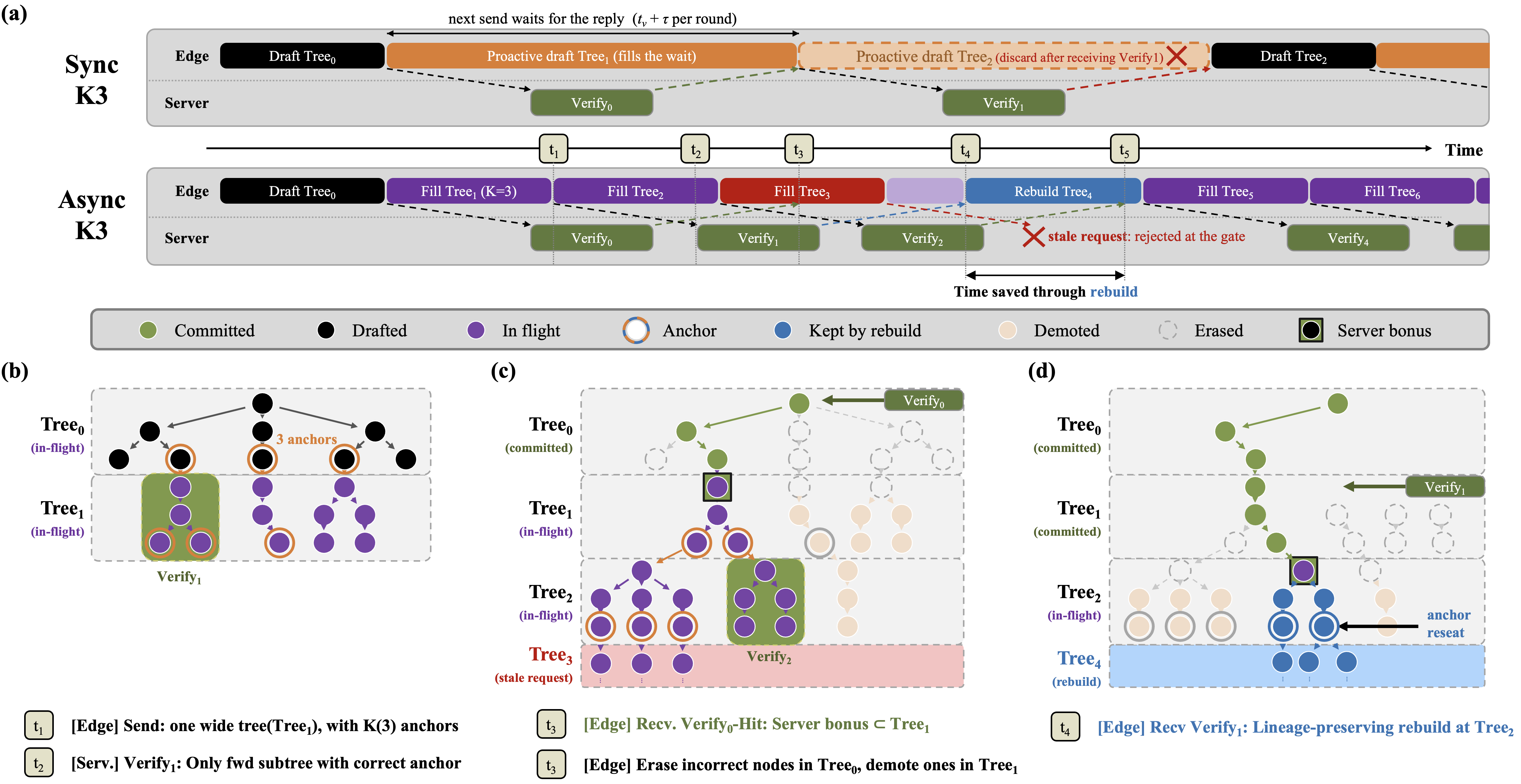}  %
   \caption{\clrev{Anchor-hedged asynchronous verification pipeline. \textbf{(a)} Edge and server timelines under the synchronous and asynchronous schedules. \textbf{(b--d)} Edge KV state of Async-K3: \textbf{(b)} $\mathrm{Tree}_1$ drafted under three anchors of $\mathrm{Tree}_0$ and in flight; \textbf{(c)} reply to $\mathrm{Tree}_0$ ($\mathrm{Verify}_0$) received; \textbf{(d)} reply to $\mathrm{Tree}_1$ ($\mathrm{Verify}_1$) received, lineage-preserving rebuild at $\mathrm{Tree}_2$.}}
  \label{fig:async_overview}
  \vspace{-0.9em}  %
\end{figure}

\subsection{\new{\clrev{Asynchronous} Verification Pipeline}}
\label{sec:pipeline}

\clrev{Under a synchronous schedule the edge issues the next verification request only after the previous reply has arrived, so the
request rate is bounded by the round-trip time. Under the asynchronous schedule the edge continues drafting from the leaves it expects to
be accepted and issues each fixed-budget tree as soon as it is complete, so several requests are in flight at any time
(Figure~\ref{fig:async_overview}a). The number of requests in flight is determined by the round-trip time and the drafting rate
rather than by a parameter, and a longer round trip is covered by additional trees rather than by deeper ones. A single configuration
(draft depth $5$, $K{=}3$; $K$ sweep in Appendix~\ref{app:ksweep}) is therefore used for all RTTs, edge GPUs and model pairs in \S\ref{sec:experiments}, whereas a synchronous
draft has to be re-tuned to the wait in each setting (\S\ref{sec:systems_results}, Appendix~\ref{app:sync_tuning}).}

\clrev{\textbf{What crosses the network.} As in SpecEdge, no KV cache, hidden state or logit is exchanged; \pfit{each side keeps
its own model's cache}, and consistency rests on token IDs and positions. A \emph{request} carries the draft tree, \pfit{its \emph{base}}
(the committed position last seen by the edge and the tokens assumed accepted beyond it) and, \pfit{for a query's first request}, the
prompt. A \emph{reply} carries the accepted path, the bonus token and \pfit{the committed position}, from which the edge
\pfit{resynchronizes on rejection}.}

\clrev{The \(j\)-th request of a session is \(r_j=(A_j,T_j)\): its \emph{base} \(A_j\), the set of \emph{anchors}---leaves of the
previous tree, transmitted as the token paths from the committed position last seen by the edge to each anchor---and its draft
tree \(T_j\), one fixed-budget draft subtree rooted at each anchor. The edge does not guess the target's bonus
token; the first level of each subtree holds the anchor's drafted candidates for that position. Each anchor is a bet, placed before
the verdict of \(r_{j-1}\) arrives, that the accepted path will end at that anchor; \(T_j\) is usable only if one of these bets turns out
to be correct. Since a tree is drafted before the verdict that fixes its place in the sequence, its coordinates---token positions and
KV-cache slots---are relative to its anchor; the server rebases the matched subtree onto its committed position, and the edge rebases
surviving drafts in the same way when it applies a verdict (\S\ref{sec:kroot}), so a hit costs neither side any redrafting or
renumbering.}

\clrev{On the server side, requests of a session are processed in the order received. When \(r_j\) reaches the front of the queue,
let \(\ell^{*}_{j-1}\) and \(z^{*}_{j-1}\) be the accepted leaf and the bonus token of the verdict on \(r_{j-1}\). Before executing the
target model, the server applies the \emph{gate}, which classifies \(r_j\) as \emph{valid} or \emph{stale}:
\[
\mathrm{valid}(r_j)\;\Longleftrightarrow\;\mathrm{valid}(r_{j-1})\;\wedge\;\ell^{*}_{j-1}\in A_j\;\wedge\;z^{*}_{j-1}\in T_j ;
\]
that is, \(r_j\) is valid if \(r_{j-1}\) was valid, one of its anchor bets was right, and the committed bonus token is in \(T_j\)
(as a child of that anchor). The server checks this by comparing the declared anchor paths position by position with \pfit{its
committed tokens}. A stale request is rejected without a target-model forward or state change (Figure~\ref{fig:async_overview}a, $t_4$--$t_5$), and
subsequent requests are rejected in turn until \pfit{a rebuilt tree resets the chain}. A valid request is verified only under its matching
anchor (Figure~\ref{fig:async_overview}b, $\mathrm{Tree}_1$); the server \pfit{commits and returns} the accepted path and the bonus token
\pfit{as the reply}.}

\clrev{The edge also applies replies in order. On each reply, it adopts the server's verdict and
updates its draft trees to match: drafts that continue the committed prefix stay valid, together with
the in-flight requests built on them, and the rest are dropped from the active tree
(\S\ref{sec:kroot} describes which drafts survive and where the anchors are re-seated); replies to requests issued before a rebuild are ignored.}

\subsection{\clrev{Continuation Hedging at Tree Boundaries}}
\label{sec:kroot}

\clrev{With a single anchor, a miss at a tree boundary invalidates every request in flight, and the longer the round trip, the more
\pfit{in-flight requests depend on each anchor}. We therefore hedge across up to \(K\) anchors: in Figure~\ref{fig:async_overview}b, \(\mathrm{Tree}_1\) is
drafted under three anchors of \(\mathrm{Tree}_0\). \pfit{The anchors are placed at the forks where the accepted path is most likely to leave the
draft's most probable path, judged by the local margin at the fork rather than by cumulative probability,
which favors the forks nearest the leaf; a subtree is grown under each anchor with an equal share of the beams, and the verdict
keeps the surviving subtree (Appendix~\ref{app:anchor_rule} gives the rule).}}

\clrev{The \(K\)-times wider tree \pfit{does not burden the link}: as in SpecEdge, a request carries only token IDs and tree structure (four
integer vectors) plus the base fields the gate needs---about $3$~KB for the ${\approx}100$-node $K{=}3$ tree, against ${\approx}1$~KB for a
$K{=}1$ tree and a ${\sim}0.1$~KB reply---so the cloud path remains latency-bound rather than bandwidth-bound. The wider draft forward
costs the edge about $10\%$ per forward; Appendix~\ref{app:fwd} shows that the fewer rejected forwards \pfit{outweigh it}.}

\clrev{\textbf{Lineage-preserving rebuild.} A verdict \emph{hits} when the accepted path ends at one of the anchors and the bonus token
is among that anchor's drafted children; drafting then simply continues under that anchor and the other hedges are demoted
(Figure~\ref{fig:async_overview}c, $\mathrm{Verify}_0$). A verdict may also leave the predicted path before its tip. In
Figure~\ref{fig:async_overview}d the verdict of $\mathrm{Verify}_1$ arrives while $\mathrm{Tree}_2$ and $\mathrm{Tree}_3$ are in
flight. The edge keeps the draft tokens that continue the accepted path together with the server's bonus token and demotes the rest,
which leaves $\mathrm{Tree}_3$ stale; it then looks back, over at most a few levels, for the deepest level of the surviving tree
that still offers \(K\) distinct continuations, re-seats the anchors there---here at the surviving leaves of $\mathrm{Tree}_2$---and
resumes drafting from them as $\mathrm{Tree}_4$, so that the hedge is re-established rather than continued from a single survivor.
Only when nothing survives does the edge draft a fresh tree from the bonus token.}

\clrev{\textbf{Synchronous variant.} The same hedge applies to a synchronous schedule---\(K\) proactive subtrees grown during the
round trip, the matching one sent after the reply---which \S\ref{sec:systems_results} uses (\sSthree) to separate the contribution of
hedging from that of asynchronous verification.}

\paragraph{\clrev{Correctness.}}
\clrev{With in-order processing per session and non-mutating rejection of stale requests, the request that advances the committed
prefix is always verified from exactly that prefix, so its verdict is what serial speculative decoding would produce; hedging, the
rebuild and the trees drafted ahead change only what the edge sends, never what the server commits. Asynchronous verification with
hedging therefore yields output of the same quality as the synchronous schedule.}

\section{\new{RTT-Conditioned Routing}}
\label{sec:router}

\new{Given query \(q\) and current RTT estimate \(\hat{\tau}\), the
router selects among \EdgeDirect{}, \EdgeCoD{},
\EdgeThink{}, and \SpecCloud{}. The first three use the
same edge model under different reasoning policies; \SpecCloud{}
uses the \clrev{anchor-hedged asynchronous pipeline (\sAthree; \S\ref{sec:backend})}. The decision
therefore jointly chooses execution placement and local test-time
compute.}

\subsection{\new{Per-Mode Estimation}}
\label{sec:router-estimation}

\new{A frozen encoder maps \(q\) to \(x=E(q)\). For each mode \(m\),
cosine-similarity \(k\)-nearest-neighbor regressors estimate correctness
\(\hat A_m(q)\) and decode length \(\hat O_m(q)\):}
{\color{newblue}\par\noindent
\begin{minipage}{0.49\linewidth}
\begin{equation}
\hat A_m(q)=\frac{1}{k_A}\sum_{i\in\mathcal N_{k_A}(x)} y_{m,i},
\label{eq:router-quality}
\end{equation}
\end{minipage}\hfill
\begin{minipage}{0.49\linewidth}
\begin{equation}
\hat O_m(q)=\frac{1}{k_O}\sum_{i\in\mathcal N_{k_O}(x)} O_{m,i},
\label{eq:router-length}
\end{equation}
\end{minipage}\par\smallskip}
\noindent\new{where \(y_{m,i}\in[0,1]\) is the correctness label and \(O_{m,i}\) is
the realized decode-token count of mode \(m\) on fitting query \(i\).
The neighborhood sizes \(k_A\) and \(k_O\) are selected independently
on validation data.}

\new{Let \(P(q)\) be the known prompt length. We convert the predicted
output length into request-completion latency \(\hat L_m(q,\hat{\tau})\)
using measured hardware profiles \(F_m\) for the edge modes and
\(F_{\mathrm{sc}}\) for the cloud path:
\begin{equation}
\hat L_m(q,\hat{\tau}) =
\begin{cases}
F_m\!\left(P(q),\hat O_m(q)\right),
    & m\in\mathcal M_{\mathrm{edge}},\\[2pt]
F_{\mathrm{sc}}\!\left(P(q),\hat O_m(q),\hat{\tau}\right),
    & m=\mathrm{sc},
\end{cases}
\label{eq:router-latency}
\end{equation}
where \(\mathcal M_{\mathrm{edge}}\) contains the three local modes and
\(\mathrm{sc}\) denotes \SpecCloud{}. Local profiles are linear in prompt
and output length, \(P(q)/\pi_m+\hat O_m(q)/\delta_m\), with per-mode
prefill throughput \(\pi_m\) and decode throughput \(\delta_m\) measured
on the deployment. The cloud profile additionally includes
both model prefills, communication, verification, pipeline fill and
drain, and reconciliation. We obtain these profiles from calibration
traces and interpolate across profiled RTTs, rather than approximating
the asynchronous schedule with a constant overlap factor.}

\subsection{\new{Routing Objective}}
\label{sec:router-objective}

\new{For trade-off parameter \(\lambda\in[0,1]\), the router selects
\begin{equation}
g_{\lambda}(q,\hat{\tau})
=
\arg\min_{m\in\mathcal M}
(1-\lambda)\bigl(1-\hat A_m(q)\bigr)
+
\lambda
\frac{\hat L_m(q,\hat{\tau})}{L_{\mathrm{ref}}},
\label{eq:router-objective}
\end{equation}
where \(L_{\mathrm{ref}}\) is the \pfit{validation-set mean completion latency}.
RTT enters only through \(F_{\mathrm{sc}}\)\pfit{, so} the same
query may use cloud verification on a fast link and local reasoning on
\pfit{a slow one}.}

\new{Operating points are selected without test labels. For each target
accuracy tolerance \(\epsilon\), we choose \(\lambda\) on validation
data to minimize mean completion latency over a prespecified RTT grid,
subject to accuracy remaining within \(\epsilon\) of the cloud target.
The matched-accuracy operating points use \(\epsilon{=}0\); varying
\(\epsilon\) traces a deployable accuracy--latency frontier.
We then freeze \(\lambda\), the neighborhood sizes, latency profiles,
and normalization constants before the test sweep. Any test-set sweep
of \(\lambda\) is reported only as an oracle frontier, not as a
deployable operating point.}

\subsection{\new{Construction and Overhead}}
\label{sec:router-construction}

\new{We partition routing data into disjoint fitting, validation, and
test sets. On the fitting split, we execute all four modes, record output
lengths and completion traces, assign correctness labels with a single
LLM judge (Qwen3.5-27B), embed the queries, and build the retrieval
index. Validation data select
\(k_A\), \(k_O\), the latency-profile parameters, \(L_{\mathrm{ref}}\),
and \(\lambda\). For out-of-distribution evaluation, the complete policy
is applied without refitting or recalibration.}

\new{At runtime, the encoder forward (a \rrev{Qwen3-Embedding-0.6B} model; larger routers add latency without routing gains, Appendix~\ref{app:router_size}) overlaps the edge model's prompt
prefill. Because this cost is common to all modes, it is omitted from
Eq.~\eqref{eq:router-objective}; any non-overlapped residual \pfit{counts}
in the reported end-to-end latency. The router requires no gradient
training and \pfit{modifies neither model}.}
\section{Experiments}
\label{sec:experiments}
\begin{table}[t]
\centering
\scriptsize
\setlength{\tabcolsep}{3pt}
\renewcommand{\arraystretch}{0.8}  %
\resizebox{\linewidth}{!}{%
\begin{tabular}{ll@{\hspace{1.4em}}ll@{\hspace{1.4em}}ll}
\toprule
\multicolumn{2}{l}{\textbf{Hardware}} & \multicolumn{2}{l}{\textbf{Network} (edge$\leftrightarrow$cloud)} & \multicolumn{2}{l}{\textbf{Models}} \\
\cmidrule(r){1-2}\cmidrule(lr){3-4}\cmidrule(l){5-6}
Edge device  & RTX 4090  & \clrev{Base round trip$^{\dagger}$} & \clrev{$12.5$~ms}      & Draft (edge)   & Qwen3-1.7B (FP16) \\
Cloud server & A100-40GB & Added latency swept                 & $0$--$500$~ms          & Target (cloud) & Qwen3-14B (FP16)  \\
             &           & \clrev{Effective RTT}               & \clrev{measured per row} & \clrev{Router modes} & \clrev{\EdgeDirect, \EdgeCoD, \EdgeThink, \SpecCloud} \\
\addlinespace[3pt]
\multicolumn{6}{l}{\clrev{\textbf{Variants}}} \\
\cmidrule{1-6}
\multicolumn{2}{l}{\clrev{\sUp: released SpecEdge}} & \multicolumn{2}{l}{\clrev{\sSone: our reproduction of \sUp{} (sync., $K{=}1$)}} & \multicolumn{2}{l}{\clrev{\sB: \sSone{} $+$ edge draft optimizations (cuDNN attention, KV-length buckets)}} \\
\multicolumn{2}{l}{\clrev{\sSthree: \sB{} $+$ hedging ($K{=}3$)}} & \multicolumn{2}{l}{\clrev{\sAone: \sB{} $+$ asynchronous verification}} & \multicolumn{2}{l}{\clrev{\sAthree: \sSthree{} $+$ asynchronous verification}} \\
\multicolumn{6}{l}{\clrev{\sR: \sAthree{} $+$ router (full system)}} \\
\bottomrule
\multicolumn{6}{l}{\clrev{$^{\dagger}$Mean network path $+$ gRPC handling on both ends, no added latency; effective RTT is measured the same way (Table~\ref{tab:systems_current}).}} \\
\end{tabular}}
\caption{\edit{Experimental setup: hardware, \pfit{models}} \clrev{and the variants that appear in the experiments.}}

\label{tab:setup}
\vspace{-0.9em}  %
\end{table}

\clrev{The evaluation proceeds at two levels. \S\ref{sec:systems_results} isolates the \SpecCloud{} path and attributes its
throughput gain across RTT to the three terms of Eq.~(\ref{eq:sync_latency}): the draft time $t_d$ (edge-draft optimizations),
the acceptance $\alpha$ (continuation hedging), and the overlap $o$ (asynchronous verification). \S\ref{sec:main_results}
evaluates the complete system, whose per-query routing decisions shift with the RTT, and reports throughput at matched
target-model accuracy together with the resulting server compute.}

\subsection{Experimental Setup}

\clrev{Table~\ref{tab:setup} summarizes the hardware, models and network setup: an RTX~4090 edge drafting with Qwen3-1.7B, an A100
cloud verifying with Qwen3-14B, and gRPC between them; Appendix~\ref{app:edge3090} repeats the sweep on an RTX~3090 edge and
Appendix~\ref{app:models} on two further model pairs (Llama-3.2-1B/3.1-8B and Qwen3-1.7B/32B). The router chooses among the four modes in Table~\ref{tab:setup}, using
per-mode latency profiles (Eq.~\eqref{eq:router-latency}) measured on this deployment. %
\pfit{The added RTT} is injected symmetrically between send and receive, from $0$ to $500$~ms, on top of the
${\approx}12.5$~ms base round trip and the server's verification time. All systems decode greedily (temperature~$0$);
Appendix~\ref{app:temperature} reports a temperature-$0.7$ spot check, and Appendix~\ref{app:details} lists the remaining run
settings.}

\paragraph{Variants.} \clrev{\sUp{} is the released SpecEdge code~\cite{park2025specedge} and \sSone{} our reproduction of it---the same
synchronous $K{=}1$ schedule with \pfit{the four correctness fixes of Appendix~\ref{sec:upstream}}---and the reference for
every speedup we report. Table~\ref{tab:setup} defines the variants that appear in the results.}

\paragraph{Datasets.} SPROUT~\cite{somerstep2025carrot} is our mixed-difficulty
routing benchmark, spanning GPQA, MuSR, MMLU-Pro, MATH, OpenHermes, and RAGBench.
OlympiadBench~\cite{he-etal-2024-olympiadbench} evaluates out-of-distribution transfer on unseen
Olympiad-level mathematics and physics without refitting or recalibration
(Appendix~\ref{sec:generalization}). %
\clrev{SpecBench~\cite{xia-etal-2024-unlocking} is the speculative-decoding benchmark SpecEdge reports on.}

\paragraph{Metrics.} \clrev{For \SpecCloud{} we report decoding throughput: the output tokens of a benchmark's $32$-query run \pfit{divided by
its client wall time}. Because every \SpecCloud{} variant emits the target model's verified output for a query (the same text up to
fp16 near-ties), a throughput ratio is also a wall-time ratio. For the full system\pfit{}, whose modes emit outputs of different
lengths, \pfit{speedup is defined on wall time: the summed} per-query completion latencies (Eq.~\ref{eq:router-latency}) of the
routed workload against running every query on \SpecCloud{} at matched target-model accuracy, \pfit{with} the accuracy--latency
frontier; \S\ref{sec:main_results} defines \pfit{the protocol}. Because many edge devices share \pfit{one server}, we also
compare the server-side runtime, the verifier's GPU time per $1$k output tokens (Figure~\ref{fig:server_cost}b--d,
Appendix~\ref{app:server_cost}).}

\subsection{Spec-Cloud Ablations and Throughput Across RTT}
\label{sec:systems_results}

\clrev{Table~\ref{tab:systems} reports, per benchmark and added latency, the decoding throughput of \sUp{} and \sSone{} in tok/s and
that of every other variant as a speedup over \sSone{}. \sB{} adds two edge-side optimizations
to \sSone, cuDNN attention and CUDA graphs captured at KV-length buckets, which shorten the draft forward from $15.6$ to $8.9$~ms on the
RTX~3090 without touching verification (Appendix~\ref{app:bucketing}); every variant to its right includes them. \sSthree{} adds
continuation hedging ($K{=}3$) to the synchronous schedule (the variant of \S\ref{sec:kroot}), \sAone{} is the asynchronous schedule
without hedging and \sAthree{} with it, so \sSthree{} and \sAone{} each isolate one step of \sAthree.}

\paragraph{Draft depth.}
\clrev{The synchronous variants follow SpecEdge's schedule but not its depth rule, verification time ${\approx}$ drafting time
$+$ RTT, which has no solution once the RTT exceeds the verification time \pfit{(from $50$~ms added latency)}. Following the
released code, they instead cap main and proactive tree alike at $d_{\max} = 3 + 0.06\cdot\mathrm{RTT}$ (Table~\ref{tab:systems}), so
that proactive drafting at \sSone's forward time spans the round trip; from $200$~ms on the $32$-node budget, not the cap, bounds
the depth (${\approx}14$ levels), and \pfit{a larger budget does not help, since it lengthens every forward}, above all the
server's (Appendix~\ref{app:edge_budget}).}

\paragraph{Re-tuned baselines.}
\clrev{Since \sB's faster forward leaves part of the wait unused under this cap, Appendix~\ref{app:depth_rule} re-tunes the
synchronous variants with a shallow main tree and a proactive tree \pfit{filling the wait}: the hit rate rises, but the proactive tree
still deepens with the RTT, the outcome varies by benchmark at $0$~ms, and \sAthree's lead holds from $50$~ms on
(Table~\ref{tab:fill}). The asynchronous schedule needs no such rule: one configuration (draft depth $5$; $K{=}1$ for \sAone, $3$ for
\sAthree, within $4\%$ of the best $K$ at every RTT, Appendix~\ref{app:ksweep}) serves every RTT, both edge GPUs and all three
model pairs, because a longer round trip is absorbed by more trees in flight rather than a deeper proactive tree.}

\paragraph{Throughput across RTT.}
\clrev{\sAthree{} is the fastest \SpecCloud{} variant on all three benchmarks at every added latency $\ge 25$~ms, up to
$1.50$/$1.45$/$1.48\times$ over \sSone{} on SPROUT/SpecBench/OlympiadBench; only at $0$~ms on SpecBench, with no round trip to hide,
does \sSthree{} lead ($1.10\times$ vs.\ $1.02\times$). The steps compound rather than add: hedging alone changes a synchronous schedule
by at most $6\%$ (\sSthree{} vs.\ \sB) and asynchronous verification alone gives $1.11$--$1.25\times$ at $\ge 100$~ms (\sAone), yet
hedging on top of it adds a further $15$--$22\%$ (\sAthree{} vs.\ \sAone), because a missed anchor costs a synchronous round one
proactive draft but invalidates every request in flight under the asynchronous schedule. Appendix~\ref{app:fwd} traces the gain to
the draft forwards: \pfit{hedged anchors and the lineage-preserving rebuild keep more of \sAthree's trees on the accepted path}, so it
spends $24$--$27\%$ less \pfit{on rejected drafting} per committed token than the re-tuned \sB{} despite a $12\%$ longer forward. Under
bursty latency the asynchronous schedule's lead widens further (Appendix~\ref{sec:jitter_ablation}).}

\providecolor{rtrrev}{RGB}{200,70,0}\providecommand{\rrev}[1]{\textcolor{rtrrev}{#1}}  %
\begin{table*}[t]
\centering
\footnotesize
\setlength{\tabcolsep}{3pt}
\renewcommand{\arraystretch}{0.9}
\color{clrevblue}
\resizebox{\textwidth}{!}{%
\begin{tabular}{r r r rr ccccc rr ccccc rr ccccc}
\toprule
& & & \multicolumn{7}{c}{\textbf{SPROUT} (1024-tok)} & \multicolumn{7}{c}{\textbf{SpecBench} (256-tok)} & \multicolumn{7}{c}{\textbf{OlympiadBench} (1024-tok)} \\
\cmidrule(lr){4-10}\cmidrule(lr){11-17}\cmidrule(lr){18-24}
Added & Eff.\ RTT & $d_{\max}$ & \sUp & \sSone & \sB & \sSthree & \sAone & \sAthree & \rrev{\texttt{\bfseries R}} & \sUp & \sSone & \sB & \sSthree & \sAone & \sAthree & \rrev{\texttt{\bfseries R}} & \sUp & \sSone & \sB & \sSthree & \sAone & \sAthree & \rrev{\texttt{\bfseries R}} \\
(ms) & (ms) & (sync) & \multicolumn{2}{c}{tok/s} & \multicolumn{5}{c}{$\times$\,\sSone} & \multicolumn{2}{c}{tok/s} & \multicolumn{5}{c}{$\times$\,\sSone} & \multicolumn{2}{c}{tok/s} & \multicolumn{5}{c}{$\times$\,\sSone} \\
\midrule
0 & 12 & 3 & 38.0 & 51.9 & 1.13 & 1.16 & 1.12 & 1.18 & \rrev{\textbf{1.24}} & 39.5 & 47.5 & 1.06 & 1.10 & 0.95 & 1.02 & \rrev{\textbf{1.12}} & 39.7 & 65.4 & 1.06 & 1.09 & 1.05 & \textbf{1.15} & \rrev{1.13} \\
25 & 38 & 5 & 30.0 & 44.8 & 1.17 & 1.20 & 1.09 & 1.25 & \rrev{\textbf{1.31}} & 30.0 & 37.8 & 1.10 & 1.10 & 0.97 & 1.12 & \rrev{\textbf{1.14}} & 32.0 & 58.7 & 1.08 & 1.13 & 1.03 & \textbf{1.17} & \rrev{1.14} \\
50 & 64 & 6 & 23.4 & 38.9 & 1.12 & 1.16 & 1.10 & 1.28 & \rrev{\textbf{1.38}} & 24.0 & 30.9 & 1.07 & 1.09 & 1.02 & 1.21 & \rrev{\textbf{1.25}} & 26.0 & 51.2 & 1.06 & 1.10 & 1.08 & \textbf{1.24} & \rrev{1.22} \\
100 & 115 & 9 & 18.8 & 29.9 & 1.13 & 1.15 & 1.18 & 1.39 & \rrev{\textbf{1.59}} & 17.4 & 22.2 & 1.09 & 1.09 & 1.11 & 1.33 & \textbf{\rrev{1.33}} & 21.8 & 41.7 & 1.08 & 1.11 & 1.13 & \textbf{1.32} & \rrev{1.30} \\
150 & 169 & 12 & 15.8 & 24.7 & 1.14 & 1.15 & 1.23 & 1.46 & \rrev{\textbf{1.71}} & 14.7 & 17.2 & 1.10 & 1.11 & 1.22 & 1.42 & \rrev{\textbf{1.45}} & 19.3 & 35.9 & 1.08 & 1.10 & 1.17 & \textbf{1.35} & \rrev{1.28} \\
200 & 218 & 15 & 14.3 & 21.5 & 1.16 & 1.15 & 1.22 & 1.49 & \rrev{\textbf{1.84}} & 12.8 & 14.5 & 1.10 & 1.12 & 1.22 & 1.45 & \rrev{\textbf{1.50}} & 17.5 & 31.4 & 1.09 & 1.13 & 1.17 & \textbf{1.41} & \rrev{1.37} \\
250 & 272 & 18 & 12.6 & 19.1 & 1.14 & 1.14 & 1.24 & 1.48 & \rrev{\textbf{1.95}} & 11.1 & 12.6 & 1.10 & 1.11 & 1.25 & 1.44 & \rrev{\textbf{1.50}} & 16.0 & 27.8 & 1.08 & 1.12 & 1.20 & 1.41 & \rrev{\textbf{1.55}} \\
300 & 321 & 21 & 12.0 & 17.2 & 1.12 & 1.15 & 1.24 & 1.48 & \rrev{\textbf{1.97}} & 10.4 & 11.4 & 1.09 & 1.10 & 1.22 & 1.42 & \rrev{\textbf{1.50}} & 14.3 & 25.6 & 1.06 & 1.10 & 1.17 & 1.41 & \rrev{\textbf{1.65}} \\
350 & 379 & 24 & 11.1 & 15.8 & 1.10 & 1.14 & 1.22 & 1.44 & \rrev{\textbf{2.02}} & 8.6 & 10.3 & 1.08 & 1.10 & 1.22 & 1.43 & \rrev{\textbf{1.50}} & 13.5 & 23.1 & 1.06 & 1.12 & 1.20 & 1.46 & \rrev{\textbf{1.78}} \\
400 & 430 & 27 & 9.8 & 14.5 & 1.11 & 1.13 & 1.24 & 1.50 & \rrev{\textbf{2.13}} & 8.5 & 9.6 & 1.06 & 1.07 & 1.20 & 1.39 & \rrev{\textbf{1.47}} & 12.7 & 21.4 & 1.06 & 1.11 & 1.23 & 1.48 & \rrev{\textbf{1.98}} \\
450 & 482 & 30 & 9.6 & 13.4 & 1.12 & 1.13 & 1.24 & 1.46 & \rrev{\textbf{2.16}} & 8.1 & 8.9 & 1.07 & 1.07 & 1.18 & 1.37 & \rrev{\textbf{1.47}} & 12.0 & 19.8 & 1.04 & 1.09 & 1.24 & 1.46 & \rrev{\textbf{2.13}} \\
500 & 535 & 32 & 8.7 & 12.5 & 1.10 & 1.11 & 1.24 & 1.45 & \rrev{\textbf{2.21}} & 7.2 & 8.3 & 1.06 & 1.06 & 1.18 & 1.36 & \rrev{\textbf{1.47}} & 11.1 & 18.6 & 1.05 & 1.08 & 1.21 & 1.45 & \rrev{\textbf{2.22}} \\
\bottomrule
\end{tabular}}
\caption{\clrev{Decoding throughput and speedup across added RTT on the RTX~4090 edge. \sUp{} and \sSone{} in tok/s; the other columns are speedups over \sSone; bold $=$ best per benchmark and RTT.}}
\label{tab:systems_current}\label{tab:systems}  %
\vspace{-1.0em}  %
\end{table*}

\subsection{Full-System Throughput at Matched Accuracy}
\label{sec:main_results}

\begin{figure}[t]
  \centering
  \begin{minipage}[t]{0.48\linewidth}\vspace{0pt}\centering
    \includegraphics[width=0.9\linewidth]{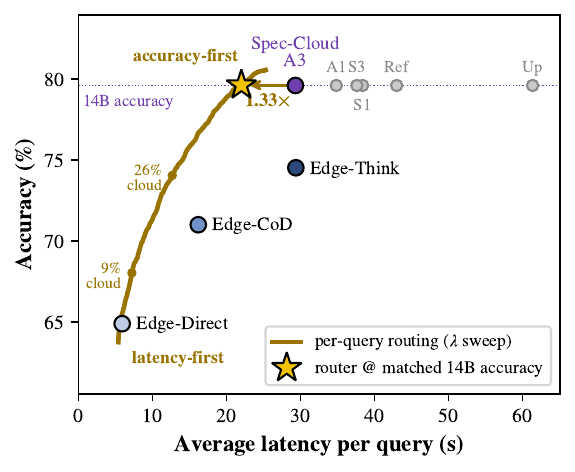}  %
    \vspace{-1.2em}
    \caption{\clrev{Accuracy vs.\ average latency per query on SPROUT at $300$~ms added latency; curve: per-query routing over $\lambda$.}}
    \label{fig:pareto}
  \end{minipage}\hfill
  \begin{minipage}[t]{0.50\linewidth}\vspace{0pt}\centering
    \begin{subfigure}[t]{0.49\linewidth}\centering\includegraphics[width=0.9\linewidth]{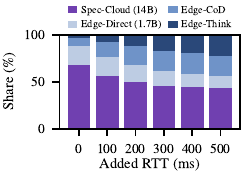}\vspace{-0.9em}\caption{}\end{subfigure}\hfill
    \begin{subfigure}[t]{0.49\linewidth}\centering\includegraphics[width=0.9\linewidth]{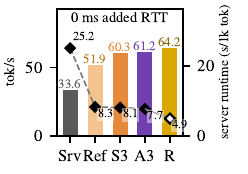}\vspace{-0.9em}\caption{}\end{subfigure}\\[0pt]
    \begin{subfigure}[t]{0.49\linewidth}\centering\includegraphics[width=0.9\linewidth]{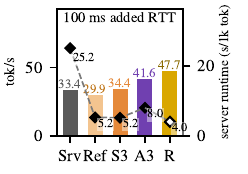}\vspace{-0.9em}\caption{}\end{subfigure}\hfill
    \begin{subfigure}[t]{0.49\linewidth}\centering\includegraphics[width=0.9\linewidth]{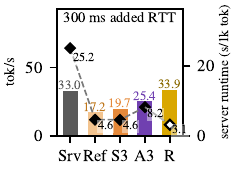}\vspace{-0.9em}\caption{}\end{subfigure}
    \vspace{-0.9em}
    \caption{\clrev{SPROUT, matched-accuracy operating point. \textbf{(a)} Routing shares vs.\ added RTT. \textbf{(b--d)} Throughput (bars) and server runtime (\(\blacklozenge\)).}}
    \label{fig:routing_shift}\label{fig:server_cost}
  \end{minipage}
  \vspace{-1.0em}  %
\end{figure}

\pfit{The \sR{} column of Table~\ref{tab:systems} reports the full system, the latency-aware router on \sAthree, as a speedup over \sSone{} at the matched target-model accuracy}
operating point: at $500$~ms \clrev{it reaches $2.21$/$1.47$/$2.22\times$ on SPROUT/SpecBench/OlympiadBench, a further
improvement over \sAthree{} that grows with the RTT.}
On SPROUT and OlympiadBench, the two benchmarks with reference answers, these
speedups do not trade quality for latency: the operating points match the always-cloud target accuracy of ${\sim}79.6\%$ and $51.6\%$, respectively.

\clrev{Figure~\ref{fig:pareto} shows the accuracy--latency plane on SPROUT at $300$~ms: each fixed mode is one point;
the router's per-query decisions trace a curve as the latency weight $\lambda$ moves from accuracy-first to latency-first \pfit{(ticks:
share of queries sent to the cloud); the star, the operating point matching the always-cloud 14B}
accuracy, \pfit{completes the workload $1.33\times$ faster than \sAthree{} alone}. The other \SpecCloud{} variants appear in
grey.}
\pfit{The max-accuracy operating points (Appendix~\ref{app:operating_points}) are complementary: on SPROUT at $300$~ms the accuracy-first end of the curve reaches $80.5\%$, $0.9$ points above the always-cloud target, at a latency still below \sAthree's, so routing can improve accuracy and wall time at once relative to running every query} on \SpecCloud.

\paragraph{Routing adaptation across RTT.}
\clrev{With the test set and the operating point held fixed, only the network latency varies. Figure~\ref{fig:routing_shift}a shows
the routing distribution on SPROUT: at zero added latency the cloud path is cheap and receives $68.2\%$ of the queries; as
the added latency rises to $500$~ms its share falls to $43.3\%$, while the combined share of \EdgeCoD{} and \EdgeThink{} rises from
$12.4\%$ to $44.0\%$, with accuracy at the always-14B $79.6\%$ throughout. The full-system gain therefore grows with the RTT because
\pfit{a growing fraction} of the queries is answered by on-device reasoning as the cloud path becomes more expensive, and cloud usage
falls accordingly. The distribution also
differs widely between benchmarks (Figure~\ref{fig:routing_grid}) and query groups (Appendix~\ref{sec:router_ablations}), so the
decisions depend on the queries as much as on the RTT; \pfit{Appendix~\ref{sec:generalization} reports out-of-distribution behavior}.}

\paragraph{\clrev{Server compute and cost.}}\label{sec:utilization}
\clrev{In realistic deployments many edge devices share one cloud server \pfit{(two-client probe: Appendix~\ref{app:edge3090_2client})}, so we also
report the verifier's GPU time per $1$k output tokens next to decoding throughput (Figure~\ref{fig:server_cost}b--d; \pfit{all RTTs and benchmarks} in Appendix~\ref{app:server_cost});
for Srv\pfit{, server-only speculative decoding on the A100,} the draft forwards count as server time, and \sR's server time \pfit{is \sAthree's per-token
server time scaled by the share of output tokens routed to the cloud}, since the edge modes use none. The synchronous variants use the
server less as the RTT grows, because their proactive tree lengthens with the wait and verifies become rare, but they decode slowly for
the same reason. \sAthree's server time per token is nearly constant across RTT: it keeps the verifier as busy at $500$~ms as at $0$~ms and turns that work into faster decoding. Routing lowers
server time at the same accuracy wherever queries can be answered on the edge: \pfit{$0.6$--$0.8\times$ \sSone{} on SPROUT across the sweep and
$0.08\times$ on OlympiadBench at $500$~ms}, whereas on SpecBench, where over $90\%$ of the queries stay on the cloud, it remains at
\sAthree's level ($1.1$--$1.2\times$ \sSone). Session rental cost per token, which also charges a slow session's idle time, is reported in
Appendix~\ref{app:server_cost} (Table~\ref{tab:utilization_current}).}

\label{sec:pipeline_highrtt}

\section{Conclusion}

\clrev{\pfit{\name{} addresses} variable RTT at two levels: \SpecCloud{} adds continuation hedging and an asynchronous,
prefix-gated verification pipeline that keep the edge drafting through the round trip, and an RTT-conditioned router decides per
query whether to \pfit{use} the cloud or answer on-device. \pfit{Over} $0$--$500$~ms \pfit{added latency}, the asynchronous schedule \pfit{raises}
decoding throughput by up to $1.5\times$ over the reproduced synchronous baseline with \pfit{one configuration}, and \pfit{a} fitted router
delivers up to $2.2\times$ speedup at matched target-model accuracy \pfit{and cuts} cloud usage by up to $2.3\times$.}

\label{pg:main-end}

\subsection*{Ethics statement}
This work studies systems efficiency of LLM inference (throughput, server usage,
and accuracy on public benchmarks). It involves no human subjects, no personal
data, and no new data release; all models and datasets are publicly available.
We do not foresee ethical concerns beyond those common to LLM serving.

\subsection*{Reproducibility statement}
The system description (\S\ref{sec:overview}--\S\ref{sec:router}) states every
mechanism at the level needed to re-implement it; the experimental protocol,
hardware, models, datasets, RTT emulation, and per-RTT draft-depth settings are
given in \S\ref{sec:experiments} and the appendix (Appendix~\ref{app:repro} onward).
The appendix further reports the per-dataset, per-RTT measurements behind every
figure. We will publicly release the source code of the edge client, the
verification server and the router, the run configurations of every variant,
the Table~\ref{tab:systems} measurements, the per-query router outputs behind
the router figures and tables, the full result logs, and the trained router.

\subsection*{AI use statement}
We used generative AI tools (large language models) to assist with code
refactoring, log analysis scripts, and editing of the manuscript text. All
AI-assisted code was reviewed and tested by the authors, and all AI-assisted
text was reviewed and revised by the authors. We did not use generative AI tools
to generate research ideas, experimental results, or figures. We take
responsibility for the final content of this work, including text, claims, and
artifacts produced with the aid of generative AI.

\label{pg:refs-start}
\bibliographystyle{iclr2027_conference}
\bibliography{custom}

\appendix

\section{Reproduction: Fixes to the Released SpecEdge Code}
\label{app:repro}\label{sec:upstream}

\clrev{\sUp{} is the released SpecEdge code, run unmodified apart from a network-delay hook in its server for the RTT sweep. On the
release's own reproduction configuration (SpecBench, $60$ queries, $256$ output tokens, initial draft depth $4$ with proactive depth
$3$, budget $32$, RTT~$0$) it reproduces Table~1 of the paper~\citep{park2025specedge}: $4.08$/$4.29$/$3.77$/$3.58$/$4.28$/$3.78$ generated tokens per verify on
the multi-turn/translation/summarization/QA/math/RAG tasks against the paper's $3.98$/$4.25$/$3.95$/$3.59$/$4.25$/$3.83$, within $0.2$
on every task.}

\clrev{\sSone{} is the same $K{=}1$ schedule and tree budget in our codebase with four fixes to the release.
\textbf{(1)} The KV-cache gather performs an in-place advanced-index copy with overlapping source and destination indices during
accepted-path compaction---on \emph{both} the edge drafter and the server verifier---which corrupts key/value entries and surfaces
as repeated tokens; we fix it with a clone-before-scatter.
\textbf{(2)} Under proactive depth-reduction the server's bonus token was committed to the prefix \emph{and} simultaneously re-added
as a tree candidate, so it was emitted twice.
\textbf{(3)} The freshly built proactive subtree's KV was overwritten during prefix compaction; we snapshot the union of accepted
and proactive slots and skip the tail-zeroing.
\textbf{(4)} Result logging was non-atomic, corrupting the throughput accounting the official metric reads.
These fixes, not further engineering, account for \sSone's higher throughput: the corrupted entries surface as repeated
tokens that the target rejects, so with the fixes in place, on the release's own configuration and the same $60$ queries,
acceptance rises from $3.90$ to $4.72$ tokens per verify and throughput from $37.4$ to $45.2$~tok/s ($+21\%$); in the
Table~\ref{tab:systems} grid \sSone{} runs $20$/$37$/$65\%$ faster than \sUp{} on SpecBench/SPROUT/OlympiadBench at RTT~$0$. With these fixes \sSone{} realizes the schedule of the SpecEdge paper as intended (synchronous $K{=}1$ drafting with
proactive expansion during the wait), so we adopt it, rather than \sUp, as the baseline against which every speedup in the paper is
measured. \sSone{} runs with a per-benchmark KV window sized to the workload ($3072$ tokens on SPROUT, $2048$ on SpecBench and
OlympiadBench) and one CUDA graph on the edge, and the server uses the same window for every variant; \sB{} and the variants built
on it allocate $4096$ on the edge and dispatch each draft step to a length bucket (Appendix~\ref{app:bucketing}).
Its verdicts follow the target's greedy decode; outputs of different tree shapes differ only where fp16 near-ties resolve
differently.}

\section{\clrev{Spec-Cloud: Ablations and Extensions}}
\label{app:speccloud}

\subsection{\clrev{Experimental Details}}
\label{app:details}
\clrev{\textbf{Runs.} Every cell of Tables~\ref{tab:systems_current}, \ref{tab:edge3090_current}, \ref{tab:llama_current}
and~\ref{tab:q32b_current} is one run of the published code with one edge client: the same $32$ prompts per benchmark, served one
after another in a fixed order, with generation capped at $1024$ output tokens on SPROUT and OlympiadBench and $256$ on SpecBench
(the SpecBench reproduction of Appendix~\ref{sec:upstream} uses $60$ prompts). Both models run in FP16 with Qwen3's thinking mode
off and greedy decoding on both sides (temperature $0$), seed $42$. The draft tree is capped at $32$ nodes and $32$ beams, with at most $16$ children per node; the synchronous variants apply the depth cap $d_{\max}$ of
Table~\ref{tab:systems_current} to the main and the proactive tree, the asynchronous variants use depth $5$ with $K{=}1$ (\sAone) or
$K{=}3$ (\sAthree). The server's KV window is $3072$ tokens on SPROUT and $2048$ on SpecBench and OlympiadBench for every variant; \sSone{} uses the same per-benchmark window on the edge with one CUDA graph, while \sB{} and every variant to its right allocate $4096$ and capture the edge draft's CUDA
graphs at KV-length buckets $512/1024/2048/4096$ and use cuDNN attention on the edge (Appendix~\ref{app:bucketing}); the
server keeps its full-width KV and default attention kernel throughout.}

\clrev{\textbf{Network.} Edge and cloud communicate over gRPC; the base round trip of $12.5$~ms (Table~\ref{tab:setup}) is the
measured network path plus gRPC handling at zero added latency. Added latency is injected on the server as a one-way delay of half
the added RTT on every receive and every send, so the physical link does not affect the sweep; the jitter runs of Appendix~\ref{sec:jitter_ablation} add a
stochastic delay on top.}

\clrev{\textbf{Server.} The verifier serves one client with a static batch of one. As in the released SpecEdge code, the target
model's prompt prefill of every prompt is computed before the run and loaded into the KV cache when the request arrives, so a run's
wall time contains the draft model's prefill and the decoding only. A keepalive kernel (a background GEMM at about $50\%$ duty, paused
while a verify forward runs) holds the server's GPU clocks up between verifies; without it the GPU governor parks the clocks while a
synchronous client waits out the round trip and the verify forward of the synchronous variants runs about $1.4\times$ slower at added
RTT $\ge 50$~ms, a penalty unrelated to the schedule and specific to our single-client setting: a server shared by several clients
would not idle long enough between verifies for its clocks to drop.}

\clrev{\textbf{Measurement.} Decoding throughput divides the output tokens of a run, counted by re-encoding the emitted text with the
model tokenizer so that every variant is counted identically, by the run's wall time on the client (first to last client log record).
Server-side runtime (Figure~\ref{fig:server_cost}, Appendix~\ref{app:server_cost}) is the sum of the verify-forward GPU times logged
by the server; the server's CPU-side request handling (parsing, mask construction, acceptance, reply) is not counted and would add
$22$--$25\%$ for the synchronous variants and $31$--$35\%$ for \sAthree, whose hedged trees need more mask work.}

\subsection{\new{Edge-Draft Acceleration via Length-Bucketed Graph Capture}}
\label{app:bucketing}
\new{Equation~\eqref{eq:sync_latency} treats the per-token edge draft time $t_d$ as a
fixed constant; on a consumer edge GPU it is not. The standard way to serve a
CUDA-graphed model under varying shape is to capture graphs at a few bucket shapes
and dispatch each step to the nearest---an established systems technique, which we do
not claim as a contribution. We simply apply it to the \emph{KV-length} axis: rather
than one graph at the configured maximum length (\clrev{$3072$ on SPROUT and $2048$ elsewhere for \sSone}), whose attention reads the full window
even when only the prefix is live, we capture a small set of graphs at length buckets
(max-len \clrev{$\in\{512,1024,2048,4096\}$}) and dispatch each draft step to the smallest covering
bucket, so the attention contraction scales with the live window. What matters is that
the base schedule leaves this \pedit{unexploited}---it treats edge draft time as
fixed---so the optimization \clrev{shortens the draft forward without touching verification, entirely
edge-side (server forward unchanged). Together with cuDNN attention, which cuts the same attention cost, it forms \sB.
Figure~\ref{fig:fwd_time} shows the effect on the RTX~3090 at $100$~ms added latency: the mean draft forward falls from $13$--$16$~ms (\sSone) to
$9$~ms, and with it the edge wall time per 1k committed tokens on every benchmark ($37\to29$~s on SPROUT, $47\to41$~s on SpecBench,
$25\to21$~s on OlympiadBench); across the RTT grid \sB{} is $1.10$--$1.17\times$ \sSone{} on SPROUT and $1.04$--$1.10\times$ on
SpecBench and OlympiadBench (Table~\ref{tab:systems_current})}.}

\subsection{\clrev{Re-tuning the Synchronous Baselines}}
\label{app:sync_tuning}
\clrev{The synchronous baselines have two knobs that might be expected to close the gap to the asynchronous schedule: the proactive depth cap (with the rule that sets it) and the tree budget. This section re-tunes both and shows that neither does: the re-tuned depth gains only at $0$~ms and \sAthree's lead holds from $50$~ms on, and a larger budget saturates, since both forwards grow with the tree and \sAthree{} stays $1.1$--$1.2\times$ above the best budget of each cell.}

\paragraph{\clrev{Proactive depth cap.}}
\label{app:depth_rule}
\clrev{A shorter draft forward lets the synchronous schedule draft a deeper proactive tree within the same round trip. The cap
$d_{\max}$ of Table~\ref{tab:systems_current} was set for \sSone's forward time, so with \sB's $9$~ms forward the capped draft ends before
the reply and the edge idles: at added $0$~ms the ${\approx}57$~ms wait outlasts the three-level draft (${\approx}27$~ms), and at $200$~ms on
the RTX~3090 \sB{} idles for $116$ of the $255$~ms wait (\sSone{} for $20$~ms). The right cap is a hardware- and RTT-specific number,
${\approx}$ wait $/$ draft-forward time (${\approx}9$~ms: $8.9$~ms on the 3090, $9.0$ on the 4090, $9.4$~ms for \sSthree's wider tree;
Table~\ref{tab:fill} lists the resulting caps).}

\clrev{Holding the main tree at depth $3$ is SpecEdge's own rule, a shallow initial tree with the wait left to proactive expansion, and
it trades acceptance for hit rate: at added $200$~ms on the RTX~4090 it raises the share of rounds whose proactive tree is confirmed
from $16$/$6$/$27\%$ to $35$/$27$/$46\%$ on SPROUT/SpecBench/OlympiadBench (\sB), while each verify accepts $26$--$29\%$ fewer tokens.}

\clrev{Table~\ref{tab:fill} sweeps this re-tuning of \sB{} and \sSthree{} over the Table~\ref{tab:systems_current} grid, with the main tree
held at depth $3$, the proactive cap filled to the wait and the budget unchanged. The net effect depends on the RTT: the re-tuning
gains only at low RTT, where the shallow main tree lets the proactive tree fill the wait; as the RTT grows, the deeper main tree of
the original cap accepts more tokens per verify, and from $300$~ms the re-tuned variants fall behind it. \sAthree's lead therefore
holds from $50$~ms on and grows from $3$--$11\%$ to $34$--$70\%$ at $300$--$500$~ms; neither re-tuning the depth nor enlarging the
budget (below) closes the gap, and both need the per-RTT, per-GPU tuning that the asynchronous schedule does without. Because the
trend reverses with the RTT, the main results keep the single cap $d_{\max}$, which needs no per-benchmark tuning.}
\begin{table}[H]
\centering\scriptsize
\color{clrevblue}
\setlength{\tabcolsep}{2.2pt}
\resizebox{\linewidth}{!}{%
\begin{tabular}{r rrr cccccc cccccc cccccc }
\toprule
& \multicolumn{3}{c}{\textbf{depth cap}} & \multicolumn{6}{c}{\textbf{SPROUT}} & \multicolumn{6}{c}{\textbf{SpecBench}} & \multicolumn{6}{c}{\textbf{OlympiadBench}} \\
\cmidrule(lr){2-4}\cmidrule(lr){5-10}\cmidrule(lr){11-16}\cmidrule(lr){17-22}
Added & Tab.~\ref{tab:systems_current} & \sB$^{\mathrm{fill}}$ & \sSthree$^{\mathrm{fill}}$ & \sB & \sB$^{\mathrm{fill}}$ & \sSthree & \sSthree$^{\mathrm{fill}}$ & \sAone & \sAthree & \sB & \sB$^{\mathrm{fill}}$ & \sSthree & \sSthree$^{\mathrm{fill}}$ & \sAone & \sAthree & \sB & \sB$^{\mathrm{fill}}$ & \sSthree & \sSthree$^{\mathrm{fill}}$ & \sAone & \sAthree \\
(ms) & $d_{\max}$ & proactive & proactive & \multicolumn{6}{c}{tok/s} & \multicolumn{6}{c}{tok/s} & \multicolumn{6}{c}{tok/s} \\
\midrule
0 & 3 & 5 & 5 & 58.6 & \textbf{64.9}$^{+}$ & 60.2 & \textbf{66.5}$^{+}$ & 58.1 & 61.2 & 50.4 & \textbf{51.0}$^{+}$ & 52.3 & \textbf{50.3} & 45.1 & 48.5 & 69.3 & \textbf{78.6}$^{+}$ & 71.3 & \textbf{78.6}$^{+}$ & 68.7 & 75.2 \\
50 & 6 & 9 & 9 & 43.6 & 45.5$^{+}$ & 45.1 & 47.7$^{+}$ & 42.8 & 49.8 & 33.1 & 33.6$^{+}$ & 33.7 & 35.6$^{+}$ & 31.5 & 37.4 & 54.3 & 59.4$^{+}$ & 56.3 & 61.6$^{+}$ & 55.3 & 63.5 \\
100 & 9 & 15 & 14 & 33.8 & 35.5$^{+}$ & 34.4 & 37.1$^{+}$ & 35.3 & 41.6 & 24.2 & 24.9$^{+}$ & 24.2 & 25.8$^{+}$ & 24.6 & 29.5 & 45.0 & 47.9$^{+}$ & 46.3 & 49.8$^{+}$ & 47.1 & 55.0 \\
150 & 12 & 21 & 20 & 28.2 & 28.1 & 28.4 & 30.2$^{+}$ & 30.4 & 36.1 & 18.9 & 19.4$^{+}$ & 19.1 & 20.2$^{+}$ & 21.0 & 24.4 & 38.8 & 39.0$^{+}$ & 39.5 & 40.9$^{+}$ & 42.0 & 48.5 \\
200 & 15 & 27 & 26 & 24.9 & 23.7 & 24.7 & 24.6 & 26.2 & 32.0 & 16.0 & 16.0$^{+}$ & 16.2 & 16.5$^{+}$ & 17.7 & 21.0 & 34.2 & 33.0 & 35.5 & 34.1 & 36.7 & 44.3 \\
300 & 21 & 32 & 32 & 19.3 & 17.6 & 19.8 & 18.6 & 21.3 & 25.5 & 12.4 & 11.8 & 12.5 & 12.1 & 13.9 & 16.2 & 27.1 & 24.1 & 28.2 & 25.5 & 30.0 & 36.1 \\
500 & 32 & 32 & 32 & 13.8 & 11.5 & 13.9 & 12.2 & 15.5 & 18.1 & 8.8 & 7.7 & 8.8 & 8.0 & 9.8 & 11.3 & 19.5 & 15.9 & 20.1 & 16.9 & 22.5 & 27.0 \\
\bottomrule
\multicolumn{22}{l}{$^{+}$ $=$ faster than the Table~\ref{tab:systems_current} cell; bold $=$ at or above \sAthree.} \\
\end{tabular}}
\caption{\clrev{Synchronous variants re-tuned to fill the wait (RTX 4090, 32 queries, tok/s): main tree at depth $3$, proactive cap $\approx$ wait $/$ draft-forward time (listed; one level lower for \sSthree), $32$-node budget unchanged. \sB, \sSthree, \sAone{} and \sAthree{} $=$ the Table~\ref{tab:systems_current} rows.}}
\label{tab:fill}
\end{table}

\paragraph{\clrev{Draft budget.}}
\label{app:edge_budget}

\clrev{The second knob is the tree budget: instead of a deeper proactive tree, the synchronous baseline could spend the wait on a bigger
one, since with more nodes per level the proactive tree hedges more continuations at the same depth. We take the
strongest synchronous variant (\sSthree) at its RTT-matched draft depth and sweep the tree budget from $32$ to $96$ nodes on both
edges (Table~\ref{tab:budget_current}). The pattern is uniform: a larger tree \emph{does}
accept more tokens per verify forward (e.g.\ $9.1\to11.1$ on SPROUT at added $200$~ms), but
both forwards grow with it: the verifier processes the whole tree in one forward ($42\to49$~ms
from budget $32$ to $96$) and each draft forward expands more nodes per level, i.e.\ more rows in its batch (forward time grows with the row count, if slowly,
Appendix~\ref{app:fwd}); the gains and costs largely cancel: the best budget raises throughput by
$9\%$ on average (at most $12\%$) on the RTX~4090, where budget $96$ is still the best, and by $4\%$ (at most $7\%$) on the RTX~3090, where
budget $64$ is the optimum and $96$ already falls back. Widening a single proactive tree
therefore saturates, whereas at the same added latency \sAthree{} (Tables~\ref{tab:systems_current}
and~\ref{tab:edge3090_current}) exceeds \sSthree{} at budget $32$ by $1.23\times$ on average ($1.11$--$1.30$) and \sSthree{} at its best budget
by $1.15\times$ ($1.09$--$1.24$) at $100$--$300$~ms, by keeping many budget-$32$ trees in flight across the round trip
instead of paying for one large one. The optimum is edge-dependent ($64$ on the RTX~3090, ${\ge}96$ on the RTX~4090), so the budget is a
per-device knob rather than a fixed constant.}

\begin{table}[H]
\centering\scriptsize
\setlength{\tabcolsep}{2.5pt}
\renewcommand{\arraystretch}{0.9}
\color{clrevblue}
\begin{tabular}{l l r r r r r r r r r r r r}
\toprule
& & \multicolumn{4}{c}{\textbf{SPROUT}} & \multicolumn{4}{c}{\textbf{SpecBench}} & \multicolumn{4}{c}{\textbf{OlympiadBench}} \\
\cmidrule(lr){3-6}\cmidrule(lr){7-10}\cmidrule(lr){11-14}
Edge & Budget & 50 & 100 & 200 & 300 & 50 & 100 & 200 & 300 & 50 & 100 & 200 & 300 \\
\midrule
\multicolumn{14}{l}{\emph{tok/s (bold = best budget per column)}} \\
RTX 4090 & 32 & \textbf{45.0} & 34.4 & 24.8 & 19.7 & 33.7 & 24.3 & 16.2 & 12.5 & \textbf{56.3} & 46.3 & 35.4 & 28.1 \\
 & 64 & 44.6 & 37.0 & 26.4 & 20.6 & \textbf{34.6} & 26.1 & 17.0 & 12.7 & 55.1 & 47.8 & 37.6 & 30.8 \\
 & 96 & 43.9 & \textbf{37.9} & \textbf{27.2} & \textbf{21.7} & 34.0 & \textbf{27.2} & \textbf{17.4} & \textbf{13.3} & 53.8 & \textbf{48.4} & \textbf{38.5} & \textbf{31.5} \\
\addlinespace
RTX 3090 & 32 & \textbf{43.6} & 34.0 & 24.5 & 19.7 & 33.2 & 23.9 & 15.8 & 12.4 & \textbf{55.4} & 45.9 & 35.2 & 28.3 \\
 & 64 & 42.8 & \textbf{35.3} & \textbf{25.4} & \textbf{20.4} & \textbf{33.3} & \textbf{25.2} & \textbf{16.3} & 12.5 & 53.6 & \textbf{46.3} & \textbf{36.3} & 30.2 \\
 & 96 & 39.4 & 33.7 & 24.5 & 20.3 & 30.3 & 23.8 & 15.9 & \textbf{12.7} & 48.2 & 43.4 & 35.7 & \textbf{30.4} \\
\midrule
\multicolumn{14}{l}{\emph{acc/verify $|$ verify ms}} \\
RTX 4090 & 32 & 7.4$|$42 & 8.3$|$42 & 9.1$|$42 & 9.3$|$42 & 5.5$|$39 & 5.8$|$39 & 6.0$|$39 & 6.0$|$39 & 9.1$|$39 & 10.8$|$39 & 12.4$|$39 & 12.7$|$39 \\
 & 64 & 7.9$|$47 & 9.3$|$47 & 10.5$|$47 & 10.7$|$47 & 6.1$|$43 & 6.5$|$43 & 6.7$|$43 & 6.7$|$43 & 9.6$|$43 & 11.8$|$43 & 14.3$|$43 & 15.3$|$43 \\
 & 96 & 8.2$|$49 & 9.9$|$49 & 11.1$|$49 & 11.8$|$49 & 6.3$|$45 & 7.0$|$45 & 7.1$|$45 & 7.2$|$45 & 9.8$|$45 & 12.3$|$45 & 15.1$|$45 & 16.4$|$45 \\
\addlinespace
RTX 3090 & 32 & 7.4$|$43 & 8.3$|$43 & 9.1$|$43 & 9.4$|$43 & 5.5$|$39 & 5.8$|$39 & 6.0$|$39 & 6.0$|$39 & 9.1$|$39 & 10.7$|$39 & 12.4$|$39 & 12.9$|$39 \\
 & 64 & 7.9$|$47 & 9.3$|$47 & 10.4$|$47 & 10.9$|$47 & 6.1$|$43 & 6.5$|$43 & 6.7$|$43 & 6.6$|$43 & 9.6$|$43 & 11.8$|$43 & 14.2$|$43 & 15.3$|$43 \\
 & 96 & 8.2$|$50 & 9.9$|$49 & 10.9$|$49 & 11.6$|$49 & 6.2$|$46 & 7.0$|$46 & 7.1$|$46 & 7.2$|$46 & 9.8$|$46 & 12.3$|$46 & 15.2$|$46 & 16.5$|$46 \\
\bottomrule
\end{tabular}
\caption{\clrev{Edge draft-budget sweep of \sSthree{} (1 client, tok/s; budget $32$ $=$ the \sSthree{} rows of Tables~\ref{tab:systems_current} and~\ref{tab:edge3090_current}; bold $=$ best budget per column). acc/verify $=$ accepted tokens per verify forward; verify ms $=$ mean server-side verify time.}}
\label{tab:budget_current}\label{tab:budget_sweep}  %
\end{table}

\subsection{\clrev{Anchor Selection and Hedged Growth}}
\label{app:anchor_rule}
\clrev{Let \(P^\star\) be the most probable root-to-leaf path of the current tree, \(\mu(u)=\log p_d(c^\star(u)\mid u)-\log p_d(c'(u)\mid u)\) the margin at
a fork \(u\in P^\star\) between its on-path child \(c^\star(u)\) and the runner-up \(c'(u)\), and \(F_{K-1}\) the \(K{-}1\) forks with the smallest
\(\mu\) among those with \(\mu<3\)~nats. The anchor set is
\[
A=\{\operatorname{tip}(P^\star)\}\;\cup\;\bigl\{\operatorname{tip}\bigl(P^\star(c'(u))\bigr):u\in F_{K-1}\bigr\},
\]
where \(P^\star(c'(u))\) is the most probable path through the runner-up branch; if fewer than \(K{-}1\) forks qualify, the next most
likely leaves are added. The margin is a local, depth-independent measure of where the draft hesitated; ranking alternatives by cumulative
probability would favor the forks nearest the leaf, since an earlier fork carries a longer runner-up path.
\pfit{Growth is local as well: the \(K\) subtrees share one draft forward per level, each receives an equal quota of the beams and ranks
candidates within itself, so a hedge subtree is not starved by the leading one (identical token IDs under different anchors remain distinct, since they represent different prefixes and KV states), and the tree may grow to \(K\) times the single-tree
budget until the verdict selects the surviving anchor, after which the other subtrees are discarded and the survivor is trimmed to one
budget. The next tree is drafted under anchors chosen on the surviving lineage in the same way, so the hedge carries across tree
boundaries. The synchronous \sSthree{} hedges with the same \(K\) but selects its anchors as the \(K\) most probable (leaf, next-token)
continuations of the main tree, since its proactive tree is rebuilt every round.}}

\subsection{\clrev{Forwards per Token: Where A3's Gain Comes From}}
\label{app:fwd}
\begin{figure}[H]
\centering
\begin{subfigure}[t]{0.19\linewidth}\centering\includegraphics[width=\linewidth]{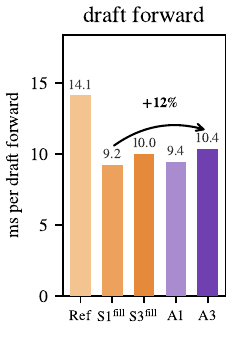}\vspace{-0.9em}\caption{}\end{subfigure}\hfill
\begin{subfigure}[t]{0.285\linewidth}\centering\includegraphics[width=\linewidth]{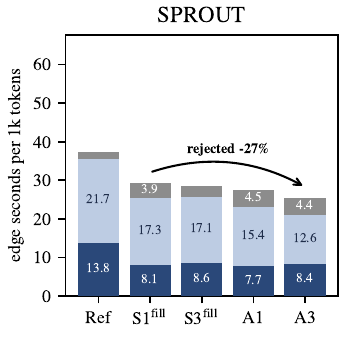}\vspace{-0.9em}\caption{}\end{subfigure}\hfill
\begin{subfigure}[t]{0.245\linewidth}\centering\includegraphics[width=\linewidth]{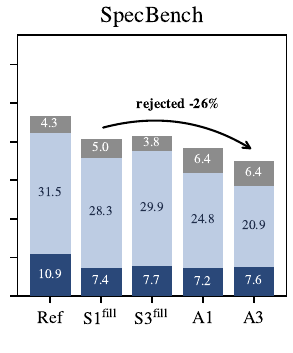}\vspace{-0.9em}\caption{}\end{subfigure}\hfill
\begin{subfigure}[t]{0.245\linewidth}\centering\includegraphics[width=\linewidth]{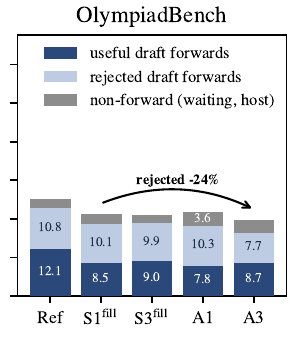}\vspace{-0.9em}\caption{}\end{subfigure}
\vspace{-0.5em}  %
\caption{\clrev{\pfit{Edge time per committed token at $100$~ms added RTT (RTX 3090). (a) Mean draft-forward time over the three benchmarks. (b--d) Edge seconds per 1k committed tokens.}}}
\label{fig:fwd_time}
\vspace{-1.0em}  %
\end{figure}

\clrev{To attribute \sAthree's gain, every draft forward on the RTX 3090 was timed with CUDA events and assigned to the tree level it
produced. A forward is \emph{useful} when
its level is on the accepted path; the count of useful forwards equals committed tokens minus bonus tokens to within $0.4\%$ in every
cell. Figure~\ref{fig:fwd_time} gives the result per committed token at $100$~ms added latency, with the synchronous
baselines in their re-tuned form (\sB$^{\mathrm{fill}}$, \sSthree$^{\mathrm{fill}}$: main tree $3$, proactive cap filled to the wait). The
draft forward itself varies little across benchmarks: $13$--$16$~ms for \sSone{} (no cuDNN attention, no buckets; \pfit{forward time follows the configured KV length: $15.8$~ms at $3072$ on SPROUT, $13.2$--$13.4$~ms at $2048$}), $9.2$ / $10.0$~ms
for \sB$^{\mathrm{fill}}$ / \sSthree$^{\mathrm{fill}}$, $9.4$ / $10.4$~ms for \sAone{} / \sAthree{}---the hedged forward carries
$\approx 30$ rows against $9$ \pfit{yet is only $12\%$ longer than \sB's, because forward time is nearly flat in the row count}. The variants differ
instead in forwards per committed token (Table~\ref{tab:fwd_count}): useful forwards are nearly equal across variants, while
rejected forwards---drafted but never accepted---are fewest for \sAthree{} on every benchmark. Filling the wait with a deeper
proactive tree does not help the synchronous edge, since the proactive tree is discarded in three rounds of four, so idle time
turns into rejected drafting at the same wall time; \sAthree's hedging instead keeps more of its in-flight trees on the accepted
lineage. Per 1k tokens the edge GPU therefore spends less time in draft forwards for \sAthree{} than for \sB$^{\mathrm{fill}}$ on
every benchmark (Figure~\ref{fig:fwd_time}b--d), with a larger share of non-forward time on SpecBench, whose short answers make
the per-query prefill and setup more visible.}
\begin{table}[H]
\centering\scriptsize
\setlength{\tabcolsep}{4pt}
\renewcommand{\arraystretch}{0.9}
\color{clrevblue}
\caption{\clrev{Draft forwards per committed token at $100$~ms added RTT (RTX 3090): useful (level on the accepted path) and rejected.}}
\label{tab:fwd_count}
\begin{tabular}{l rr rr rr }
\toprule
\textbf{Variant} & \multicolumn{2}{c}{\textbf{SPROUT}} & \multicolumn{2}{c}{\textbf{SpecBench}} & \multicolumn{2}{c}{\textbf{OlympiadBench}} \\
\cmidrule(lr){2-3} \cmidrule(lr){4-5} \cmidrule(lr){6-7}
 & useful & rejected & useful & rejected & useful & rejected \\
\midrule
\sSone & 0.88 & 1.38 & 0.83 & 2.39 & 0.90 & 0.81 \\
\sB$^{\mathrm{fill}}$ & 0.88 & 1.85 & 0.83 & 3.13 & 0.91 & 1.08 \\
\sSthree$^{\mathrm{fill}}$ & 0.88 & 1.66 & 0.83 & 2.98 & 0.91 & 0.98 \\
\sAone & 0.81 & 1.62 & 0.79 & 2.66 & 0.82 & 1.08 \\
\sAthree & 0.81 & 1.19 & 0.78 & 2.06 & 0.82 & 0.72 \\
\bottomrule
\end{tabular}
\end{table}

\subsection{\clrev{Sweep over the Number of Anchors}}  %
\label{app:ksweep}
\clrev{Table~\ref{tab:ksweep_current} sweeps the number of anchors $K$ of the asynchronous schedule on both edges. Adding anchors has two
opposing effects. Each additional anchor raises the chance that a verdict lands on a continuation already in flight, so more replies are
hits and fewer trees are rebuilt from scratch; this gain grows with the RTT, since more trees ride on each bet. But every anchor
carries its own fixed-budget subtree, so the draft forward widens with $K$ (${\approx}30$ rows per level for $K{=}3$ against ${\approx}9$
for $K{=}1$, Appendix~\ref{app:fwd}) and its cost rises, while each further anchor is placed on a less likely continuation and returns
less. The two effects cross at $K{=}3$: it is within a few percent of the best $K$ at every RTT, whereas more anchors cost throughput
at low RTT and add almost nothing at high RTT.}
\begin{table}[H]
\centering\scriptsize
\setlength{\tabcolsep}{2pt}
\renewcommand{\arraystretch}{0.9}
\color{clrevblue}
\resizebox{\linewidth}{!}{%
\begin{tabular}{l r rrrrr rrrrr rrrrr}
\toprule
\multicolumn{2}{l}{\textbf{tok/s}} & \multicolumn{5}{c}{\textbf{SPROUT}} & \multicolumn{5}{c}{\textbf{SpecBench}} & \multicolumn{5}{c}{\textbf{OlympiadBench}} \\
\cmidrule(lr){3-7} \cmidrule(lr){8-12} \cmidrule(lr){13-17}
Edge & Added (ms) & $K{=}1$ & $K{=}2$ & $K{=}3$ & $K{=}4$ & $K{=}5$ & $K{=}1$ & $K{=}2$ & $K{=}3$ & $K{=}4$ & $K{=}5$ & $K{=}1$ & $K{=}2$ & $K{=}3$ & $K{=}4$ & $K{=}5$ \\
\midrule
RTX 4090 & 0 & 58.2 & \textbf{62.0} & 61.2 & 57.9 & 56.3 & 45.0 & \textbf{48.5} & 48.3 & 45.7 & 46.5 & 68.6 & 74.6 & \textbf{74.9} & 69.9 & 67.3 \\
 & 50 & 42.6 & 49.5 & \textbf{49.9} & 47.8 & 46.9 & 31.5 & 35.5 & \textbf{37.4} & 35.9 & 36.0 & 55.1 & 63.2 & \textbf{63.6} & 59.9 & 58.5 \\
 & 100 & 35.3 & 40.2 & \textbf{41.6} & 39.0 & 36.6 & 24.7 & 28.2 & \textbf{29.6} & 29.4 & 28.6 & 47.1 & \textbf{55.2} & 55.0 & 52.4 & 51.5 \\
 & 200 & 26.2 & 30.3 & \textbf{32.1} & 31.0 & 30.7 & 17.7 & 20.0 & 21.0 & \textbf{21.1} & \textbf{21.1} & 36.8 & 43.3 & \textbf{44.3} & 42.1 & 41.1 \\
 & 300 & 21.3 & 24.0 & \textbf{25.4} & 25.3 & 25.3 & 13.9 & 15.6 & 16.2 & \textbf{16.4} & \textbf{16.4} & 30.0 & 35.3 & \textbf{36.2} & 35.7 & 35.2 \\
 & 400 & 18.0 & 20.1 & \textbf{21.8} & 21.0 & 20.7 & 11.5 & 12.7 & 13.3 & 13.5 & \textbf{13.7} & 26.3 & 29.5 & \textbf{31.6} & 30.1 & 29.6 \\
 & 500 & 15.5 & 17.3 & 18.1 & 18.0 & \textbf{18.3} & 9.8 & 10.7 & 11.3 & 11.4 & \textbf{11.5} & 22.5 & 25.6 & \textbf{26.9} & 26.5 & 26.6 \\
\addlinespace
RTX 3090 & 0 & 54.0 & 54.9 & \textbf{55.0} & 53.0 & 52.6 & 43.8 & 44.3 & \textbf{44.9} & 43.6 & 43.0 & 63.7 & \textbf{65.6} & 65.0 & 63.1 & 62.0 \\
 & 50 & 43.2 & 45.6 & \textbf{45.7} & 44.5 & 44.6 & 32.2 & 34.0 & \textbf{34.9} & 34.5 & 34.6 & 54.9 & \textbf{57.9} & 57.3 & 55.8 & 54.8 \\
 & 100 & 36.0 & 37.9 & \textbf{38.8} & 36.1 & 35.8 & 25.4 & 26.9 & 28.0 & \textbf{28.1} & \textbf{28.1} & 46.5 & 49.9 & \textbf{50.8} & 49.2 & 48.5 \\
 & 200 & 26.7 & 29.2 & \textbf{29.8} & 29.7 & 29.2 & 18.2 & 19.5 & 20.2 & 20.4 & \textbf{20.7} & 36.7 & 40.0 & \textbf{41.0} & 40.2 & 40.0 \\
 & 300 & 21.2 & 23.6 & 24.5 & 23.9 & \textbf{24.7} & 14.1 & 15.2 & 15.8 & 16.0 & \textbf{16.2} & 30.1 & 32.8 & \textbf{34.3} & 34.0 & 34.1 \\
 & 400 & 18.0 & 19.7 & \textbf{20.7} & 20.5 & 20.4 & 11.6 & 12.4 & 13.1 & 13.3 & \textbf{13.4} & 25.9 & 28.8 & 29.2 & \textbf{30.0} & 29.0 \\
 & 500 & 15.8 & 17.0 & 17.9 & \textbf{18.0} & 17.5 & 9.8 & 10.5 & 11.0 & 11.2 & \textbf{11.4} & 22.4 & 24.9 & \textbf{26.1} & 25.9 & 25.7 \\
\bottomrule
\multicolumn{17}{l}{bold $=$ best $K$ per row and benchmark.} \\
\end{tabular}}
\caption{\clrev{Sweep over the number of anchors $K$ of the asynchronous schedule.}}
\label{tab:ksweep_current}
\end{table}

\subsection{\edit{Jitter Resilience at Added RTT~0}}
\label{sec:jitter_ablation}

\clrev{The steady-state crossover (\S\ref{sec:systems_results}) understates the asynchronous schedule's value under \emph{bursty}
links. We isolate this at added RTT~0 and inject server-side latency jitter of increasing intensity under two models, a de-meaned
heavy-tailed Pareto ($\alpha{=}1.5$) AR(1) process (scale $s$) and a Gilbert--Elliott on/off burst process (mean burst $f$~ms). The
proactive baseline sizes its draft to the \emph{expected} round trip (here ${\approx}0$), so a latency spike leaves it idle; the
asynchronous schedule keeps growing and sending draft trees while a reply is delayed, and is therefore expected to degrade more
gracefully.}

\clrev{Table~\ref{tab:jitter_current} runs this sweep on both edges with the bucketed synchronous baseline (\sB), the synchronous $K{=}3$ baseline (\sSthree) and the two asynchronous schedules (\sAone, \sAthree), and \sSthree$^{\mathrm{until}}$, a variant of \sSthree{} that keeps growing the proactive tree level by level until the reply arrives, finishes the level in flight, applies the verdict and commits.
At zero jitter the asynchronous variants are at or below parity with the baselines; as the spikes intensify the ordering reverses, and \sAthree's margin over \sB{} grows with the jitter, to up to $1.56\times$ on the RTX~4090 and $1.39\times$ on the RTX~3090 at the strongest bursts, with \sAone{} following the same curve at a smaller margin.
\sSthree{} tracks \sB{} at every jitter level, so hedging alone does not hide latency spikes; the gain comes from asynchrony.
\sSthree$^{\mathrm{until}}$ recovers part of the gap on SPROUT and OlympiadBench, $1.03$--$1.28\times$ \sSthree{} at the same level and most at the strongest bursts, but not on SpecBench, whose short answers leave little wait to fill ($0.95$--$1.07\times$ \sSthree). \sAthree{} stays ahead wherever the jitter is substantial, by up to $1.23\times$ at the strongest bursts; at the two mildest levels the two tie within $\pm4\%$, and in the jitter-free control \sSthree$^{\mathrm{until}}$ edges ahead of \sAthree{} on the long-output benchmarks ($1.03$--$1.07\times$), where the proactive depth settles at $4$--$7$ and no reply is late; with a steady added RTT the gap reopens, as the filled-depth sweep of Table~\ref{tab:fill} shows (\sAthree{} ahead from $50$~ms on). A synchronous edge cannot react to a delayed reply: even \sSthree$^{\mathrm{until}}$ can only deepen one tree within the fixed budget while it waits, and commits only after the reply, so the remaining gap is structural rather than a tuning artifact.}
\begin{table}[H]
\centering\scriptsize
\setlength{\tabcolsep}{3pt}
\renewcommand{\arraystretch}{0.9}
\color{clrevblue}
\resizebox{\textwidth}{!}{%
\begin{tabular}{l ccccc c ccccc c ccccc c}
\toprule
& \multicolumn{6}{c}{\textbf{SPROUT}} & \multicolumn{6}{c}{\textbf{SpecBench}} & \multicolumn{6}{c}{\textbf{OlympiadBench}} \\
\cmidrule(lr){2-7} \cmidrule(lr){8-13} \cmidrule(lr){14-19}
Jitter (added RTT 0) & \sB & \sSthree & \sSthree$^{\mathrm{until}}$ & \sAone & \sAthree & \sAthree$/$\sB & \sB & \sSthree & \sSthree$^{\mathrm{until}}$ & \sAone & \sAthree & \sAthree$/$\sB & \sB & \sSthree & \sSthree$^{\mathrm{until}}$ & \sAone & \sAthree & \sAthree$/$\sB \\
\midrule
\multicolumn{19}{l}{\emph{RTX 4090 edge}} \\
none (control) & 58.9 & 60.0 & 63.7 & 57.4 & 61.7 & 1.05 & 50.5 & 51.2 & 48.6 & 46.0 & 49.4 & 0.98 & 70.1 & 71.3 & 78.2 & 70.4 & 74.3 & 1.06 \\
\addlinespace
Pareto-AR(1), $s{=}10$ & 49.8 & 50.7 & 55.7 & 49.1 & 57.0 & 1.14 & 39.8 & 41.2 & 40.3 & 38.5 & 41.9 & 1.05 & 59.3 & 60.5 & 69.7 & 62.6 & 69.8 & 1.18 \\
Pareto-AR(1), $s{=}30$ & 38.7 & 39.2 & 43.7 & 43.7 & 49.5 & 1.28 & 29.2 & 29.6 & 29.3 & 30.0 & 34.3 & 1.17 & 46.3 & 47.0 & 56.8 & 53.4 & 59.5 & 1.29 \\
Pareto-AR(1), $s{=}60$ & 29.0 & 29.4 & 33.4 & 36.9 & 41.1 & 1.42 & 20.4 & 20.7 & 21.0 & 24.4 & 26.4 & 1.29 & 34.8 & 35.3 & 43.2 & 44.0 & 46.3 & 1.33 \\
\addlinespace
Gilbert, $f{=}50$ & 47.6 & 48.2 & 56.2 & 50.0 & 57.3 & 1.20 & 42.3 & 43.1 & 43.3 & 40.2 & 43.8 & 1.04 & 55.6 & 56.7 & 69.9 & 61.8 & 69.7 & 1.25 \\
Gilbert, $f{=}100$ & 41.1 & 42.2 & 49.3 & 46.9 & 53.3 & 1.30 & 38.6 & 39.2 & 40.2 & 37.6 & 42.3 & 1.10 & 48.0 & 48.9 & 62.0 & 57.3 & 64.8 & 1.35 \\
Gilbert, $f{=}200$ & 33.0 & 33.6 & 40.8 & 42.8 & 48.4 & 1.47 & 32.9 & 32.9 & 35.3 & 34.9 & 39.7 & 1.21 & 37.9 & 38.7 & 49.5 & 52.4 & 59.3 & 1.56 \\
\midrule
\multicolumn{19}{l}{\emph{RTX 3090 edge}} \\
none (control) & 57.8 & 59.0 & 58.8 & 54.0 & 55.0 & 0.95 & 50.3 & 50.7 & 46.2 & 43.9 & 44.9 & 0.89 & 68.2 & 69.0 & 70.0 & 64.3 & 65.4 & 0.96 \\
\addlinespace
Pareto-AR(1), $s{=}10$ & 49.6 & 50.6 & 52.0 & 49.4 & 51.4 & 1.04 & 40.8 & 40.9 & 38.9 & 39.1 & 40.1 & 0.98 & 58.7 & 59.4 & 63.5 & 60.2 & 61.2 & 1.04 \\
Pareto-AR(1), $s{=}30$ & 39.0 & 39.7 & 41.9 & 42.4 & 45.1 & 1.16 & 29.7 & 29.9 & 29.1 & 31.0 & 30.9 & 1.04 & 46.2 & 47.1 & 53.0 & 52.0 & 54.0 & 1.17 \\
Pareto-AR(1), $s{=}60$ & 29.3 & 29.9 & 32.4 & 35.7 & 36.6 & 1.25 & 21.0 & 21.2 & 21.0 & 24.3 & 23.1 & 1.10 & 35.0 & 35.8 & 41.9 & 42.0 & 44.2 & 1.26 \\
\addlinespace
Gilbert, $f{=}50$ & 47.7 & 48.4 & 51.2 & 49.2 & 51.4 & 1.08 & 42.9 & 43.3 & 41.7 & 40.4 & 41.9 & 0.98 & 55.3 & 56.3 & 63.6 & 58.9 & 62.6 & 1.13 \\
Gilbert, $f{=}100$ & 41.4 & 42.6 & 46.9 & 46.8 & 49.0 & 1.18 & 39.2 & 39.5 & 38.7 & 39.2 & 39.7 & 1.01 & 48.2 & 49.4 & 56.6 & 55.9 & 58.1 & 1.21 \\
Gilbert, $f{=}200$ & 33.1 & 34.2 & 38.3 & -- & 44.0 & 1.33 & 33.1 & 33.5 & 34.0 & 36.7 & 37.4 & 1.13 & 38.3 & 39.2 & 46.5 & 49.8 & 53.3 & 1.39 \\
\bottomrule
\end{tabular}}
\caption{\clrev{Jitter resilience at added RTT~0 (1 client, tok/s): server-side de-meaned Pareto-AR(1) delay (scale $s$) and Gilbert--Elliott bursts (mean bad-state delay $f$~ms). \sB{}/\sSthree{} at the nominal depth $3$; the blank cell hit a deterministic $K{=}1$ livelock in both attempts.}}
\label{tab:jitter_current}\label{tab:jitter}  %
\end{table}

\subsection{\clrev{Temperature Spot Check}}
\label{app:temperature}
\clrev{All throughput results in this paper use greedy decoding (temperature~$0$),
whereas SpecEdge's original evaluation samples at temperature~$0.7$. Table~\ref{tab:temperature}
repeats three variants on SpecBench at added RTT $0$ and $300$~ms with the verifier sampling at
temperature~$0.7$. Throughput changes by $-2$ to $-11\%$ and the ordering of the variants is
unchanged on both edges, so temperature is not a confound for the comparisons above. One caveat remains:
the asynchronous schedule predicts the next committed position assuming the verifier's bonus token
is its argmax; under sampling we observed one stalled run (RTX~4090, added RTT $300$~ms) that a
retry cleared, so deployment at temperature ${>}0$ requires an explicit cursor re-synchronization step.}

\begin{table}[H]
\centering\scriptsize
\setlength{\tabcolsep}{4pt}
\renewcommand{\arraystretch}{0.9}
\color{clrevblue}
\begin{tabular}{l l cc cc}
\toprule
\multicolumn{2}{l}{\textbf{Throughput (tok/s)}} & \multicolumn{2}{c}{\textbf{Added RTT 0~ms}} & \multicolumn{2}{c}{\textbf{Added RTT 300~ms}} \\
\cmidrule(lr){3-4}\cmidrule(lr){5-6}
\textbf{Edge} & \textbf{Variant} & $T{=}0$ & $T{=}0.7$ & $T{=}0$ & $T{=}0.7$ \\
\midrule
RTX 4090 & \sSone   & 47.5 & 44.9 & 11.4 & 11.0 \\
         & \sSthree & 52.1 & 49.9 & 12.5 & 11.1 \\
         & \sAthree & 48.3 & 45.7 & 16.2 & 15.3 \\
\addlinespace
RTX 3090 & \sSone   & 41.4 & 40.4 & 10.9 & 10.5 \\
         & \sSthree & 50.7 & 48.3 & 12.4 & 11.4 \\
         & \sAthree & 44.9 & 42.3 & 15.8 & 15.4 \\
\bottomrule
\end{tabular}
\caption{\clrev{Verifier temperature $0$ vs.\ $0.7$ on SpecBench (1 client, client-span tok/s).}}
\label{tab:temperature}
\end{table}

\subsection{\clrev{Other Model Pairs}}\label{app:models}
\clrev{Two further model pairs test whether the \SpecCloud{} results are specific to Qwen3-1.7B/14B. Table~\ref{tab:llama_current} swaps the
family (Llama-3.2-1B-Instruct on the RTX~4090 edge drafting for Llama-3.1-8B-Instruct on the same A100), and Table~\ref{tab:q32b_current} enlarges the target
(Qwen3-1.7B on the edge drafting for Qwen3-32B on an H100-80GB server). In both cases the synchronous depths were re-tuned for the pair, since the $14$B depths would otherwise leave the draft idle
for part of the round trip: the 1B draft forward is ${\approx}40\%$ faster
and the $32$B verify takes $60$~ms instead of $39$~ms, so the $14$B depths fill only two thirds of its window;
the asynchronous schedule was run unchanged. With the $8$B target the crossover lies between $50$ and $100$~ms, because its $24$~ms
verify leaves less to hide. The ordering matches Table~\ref{tab:systems_current}: at zero added
latency the schedules are within $6\%$ of each other, and from $50$--$100$~ms of added latency on \sAthree{} leads with a margin that grows with the round trip.}
\begin{table}[H]
\centering\scriptsize
\setlength{\tabcolsep}{3pt}
\renewcommand{\arraystretch}{0.9}
\color{clrevblue}
\resizebox{\linewidth}{!}{%
\begin{tabular}{r r ccccc c ccccc c ccccc c}
\toprule
\multicolumn{2}{l}{\shortstack[l]{\textbf{Llama-3.2-1B}\\ \textbf{$\to$ 3.1-8B} (tok/s)}} & \multicolumn{6}{c}{\textbf{SPROUT}} & \multicolumn{6}{c}{\textbf{SpecBench}} & \multicolumn{6}{c}{\textbf{OlympiadBench}} \\
\cmidrule(lr){3-8} \cmidrule(lr){9-14} \cmidrule(lr){15-20}
\shortstack[r]{Added\\(ms)} & $d_{\max}$ & \sSone & \sB & \sSthree & \sAone & \sAthree & \sAthree$/$\sB & \sSone & \sB & \sSthree & \sAone & \sAthree & \sAthree$/$\sB & \sSone & \sB & \sSthree & \sAone & \sAthree & \sAthree$/$\sB \\
\midrule
0 & 5 & 75.7 & 88.4 & \textbf{91.3} & 75.4 & 87.7 & 0.99 & 68.2 & 73.4 & \textbf{74.5} & 62.0 & 70.0 & 0.95 & 95.6 & 101.9 & \textbf{104.5} & 88.5 & 99.1 & 0.97 \\
50 & 10 & 55.6 & 63.6 & \textbf{66.7} & 53.8 & 66.0 & 1.04 & 46.1 & 48.7 & 49.3 & 42.2 & \textbf{52.0} & 1.07 & 72.4 & 78.3 & \textbf{79.1} & 65.8 & 78.4 & 1.00 \\
100 & 15 & 40.8 & 46.1 & 46.8 & 45.1 & \textbf{53.9} & 1.17 & 33.6 & 34.8 & 35.6 & 33.0 & \textbf{40.7} & 1.17 & 53.2 & 56.3 & 58.5 & 54.6 & \textbf{64.7} & 1.15 \\
200 & 21 & 28.6 & 31.2 & 32.4 & 31.8 & \textbf{39.1} & 1.25 & 22.6 & 23.3 & 23.6 & 22.8 & \textbf{27.4} & 1.18 & 35.7 & 37.5 & 38.5 & 40.2 & \textbf{48.9} & 1.30 \\
300 & 27 & 22.1 & 24.0 & 24.6 & 25.3 & \textbf{30.1} & 1.25 & 17.1 & 17.6 & 17.8 & 17.8 & \textbf{21.3} & 1.21 & 27.4 & 28.7 & 28.8 & 32.3 & \textbf{38.2} & 1.33 \\
\bottomrule
\multicolumn{20}{l}{bold $=$ best per benchmark and RTT.} \\
\end{tabular}}
\caption{\clrev{Llama-3.2-1B-Instruct (edge draft) with Llama-3.1-8B-Instruct (cloud target); RTX 4090 edge, A100 server, 1 client, tok/s (Llama tokenizer). $d_{\max}$ $=$ synchronous depth cap re-tuned for the pair.}}
\label{tab:llama_current}
\end{table}

\begin{table}[H]
\centering\scriptsize
\setlength{\tabcolsep}{3pt}
\renewcommand{\arraystretch}{0.9}
\color{clrevblue}
\resizebox{\linewidth}{!}{%
\begin{tabular}{r r ccccc c ccccc c ccccc c}
\toprule
\multicolumn{2}{l}{\shortstack[l]{\textbf{Qwen3-1.7B}\\ \textbf{$\to$ 32B} (tok/s)}} & \multicolumn{6}{c}{\textbf{SPROUT}} & \multicolumn{6}{c}{\textbf{SpecBench}} & \multicolumn{6}{c}{\textbf{OlympiadBench}} \\
\cmidrule(lr){3-8} \cmidrule(lr){9-14} \cmidrule(lr){15-20}
\shortstack[r]{Added\\(ms)} & $d_{\max}$ & \sSone & \sB & \sSthree & \sAone & \sAthree & \sAthree$/$\sB & \sSone & \sB & \sSthree & \sAone & \sAthree & \sAthree$/$\sB & \sSone & \sB & \sSthree & \sAone & \sAthree & \sAthree$/$\sB \\
\midrule
0 & 5 & 42.8 & 51.8 & \textbf{53.0} & 46.9 & 51.0 & 0.98 & 37.3 & 41.6 & \textbf{41.7} & 36.8 & 40.6 & 0.98 & 60.9 & 67.6 & \textbf{68.1} & 61.5 & 68.0 & 1.01 \\
50 & 8 & 32.3 & 36.8 & 37.7 & 37.5 & \textbf{41.6} & 1.13 & 24.8 & 26.8 & 26.9 & 26.0 & \textbf{30.2} & 1.13 & 47.8 & 51.3 & 53.1 & 49.6 & \textbf{58.1} & 1.13 \\
100 & 13 & 24.3 & 28.6 & 30.1 & 30.3 & \textbf{35.3} & 1.23 & 17.6 & 19.5 & 19.6 & 21.0 & \textbf{24.6} & 1.26 & 40.6 & 44.6 & 45.2 & 42.1 & \textbf{51.0} & 1.14 \\
200 & 20 & 18.7 & 21.6 & 22.2 & 23.2 & \textbf{26.8} & 1.24 & 12.6 & 13.7 & 13.8 & 15.4 & \textbf{18.1} & 1.32 & 30.4 & 33.1 & 34.0 & 34.0 & \textbf{40.4} & 1.22 \\
300 & 27 & 15.5 & 17.1 & 17.2 & 18.9 & \textbf{21.6} & 1.26 & 10.1 & 10.9 & 11.0 & 12.4 & \textbf{14.1} & 1.29 & 24.8 & 26.3 & 27.1 & 28.1 & \textbf{33.6} & 1.28 \\
\bottomrule
\multicolumn{20}{l}{bold $=$ best per benchmark and RTT.} \\
\end{tabular}}
\caption{\clrev{Qwen3-1.7B (edge draft) with Qwen3-32B (cloud target) on an H100 server; RTX 4090 edge, 1 client, tok/s. $d_{\max}$ $=$ synchronous depth cap re-tuned for the pair.}}
\label{tab:q32b_current}
\end{table}

\subsection{\edit{Replication on a Second Consumer Edge GPU}}
\label{app:edge3090}

\clrev{To test whether the behavior is a property of the schedule rather than of one edge
accelerator, we repeated the current-state grid of Table~\ref{tab:systems_current} with an RTX~3090 as the edge GPU in place of
the RTX~4090, against the same A100 server, models and
configurations, at added RTT $0$--$500$~ms (Table~\ref{tab:edge3090_current}). The pattern
replicates: \sB{} and \sSthree{} gain over \sSone{}, and \sAthree{} trails the best synchronous variant at zero added latency, crosses
over at $50$~ms and leads by up to $1.3\times$ at $500$~ms on all three benchmarks, with \sAone{} between the two. Absolute
throughput is lower on the slower drafter at low RTT, and the
difference vanishes at $300$--$500$~ms, where the network hides the edge GPU.}

\begin{table}[H]
\centering\scriptsize
\color{clrevblue}
\setlength{\tabcolsep}{2.2pt}
\resizebox{\textwidth}{!}{%
\begin{tabular}{ll *{7}{r} @{\hspace{5pt}} *{7}{r} @{\hspace{5pt}} *{7}{r}}
\toprule
& & \multicolumn{7}{c}{\textbf{SPROUT}} & \multicolumn{7}{c}{\textbf{SpecBench}} & \multicolumn{7}{c}{\textbf{OlympiadBench}} \\
\cmidrule(lr){3-9}\cmidrule(lr){10-16}\cmidrule(lr){17-23}
\multicolumn{2}{l}{Added RTT (ms)} & 0 & 50 & 100 & 200 & 300 & 400 & 500 & 0 & 50 & 100 & 200 & 300 & 400 & 500 & 0 & 50 & 100 & 200 & 300 & 400 & 500 \\
\midrule
 \multirow{5}{*}{\rotatebox[origin=c]{90}{\textbf{1 client}}} & \sSone & 42.5 & 32.4 & 26.3 & 19.5 & 16.3 & 13.9 & 12.2 & 41.4 & 28.5 & 20.8 & 13.7 & 10.9 & 9.2 & 8.0 & 55.3 & 46.6 & 39.7 & 29.6 & 24.7 & 20.6 & 17.8 \\
  & \sB & 58.0 & 42.6 & 33.6 & 24.1 & 19.1 & 15.9 & 13.7 & 50.2 & 32.6 & 23.9 & 15.5 & 12.2 & 10.2 & 8.7 & 67.7 & 53.8 & 44.7 & 33.5 & 27.4 & 22.7 & 19.3 \\
  & \sSthree & \textbf{59.0} & 43.6 & 34.0 & 24.5 & 19.7 & 16.4 & 14.0 & \textbf{50.7} & 33.2 & 23.9 & 15.8 & 12.4 & 10.1 & 8.7 & \textbf{68.9} & 55.4 & 45.9 & 35.2 & 28.3 & 23.6 & 20.1 \\
  & \sAone & 54.0 & 43.2 & 36.0 & 26.7 & 21.2 & 18.0 & 15.8 & 43.8 & 32.2 & 25.4 & 18.2 & 14.1 & 11.6 & 9.8 & 63.7 & 54.9 & 46.5 & 36.7 & 30.1 & 25.9 & 22.4 \\
  & \sAthree & 55.0 & \textbf{45.7} & \textbf{38.8} & \textbf{29.8} & \textbf{24.5} & \textbf{20.7} & \textbf{17.9} & 44.9 & \textbf{34.9} & \textbf{28.0} & \textbf{20.2} & \textbf{15.8} & \textbf{13.1} & \textbf{11.0} & 65.0 & \textbf{57.3} & \textbf{50.8} & \textbf{41.0} & \textbf{34.3} & \textbf{29.2} & \textbf{26.1} \\
\midrule
 \multirow{5}{*}{\rotatebox[origin=c]{90}{\textbf{2 clients}}} & \sSone & 85.3 & 64.8 & 52.2 & 39.1 & 32.4 & 27.7 & 24.4 & 82.3 & 56.9 & 41.5 & 27.2 & 21.7 & 18.2 & 15.9 & 110.1 & 93.2 & 78.4 & 58.8 & 49.1 & 41.0 & 35.6 \\
  & \sB & 94.5 & 84.2 & 66.0 & 47.5 & 37.9 & 31.8 & 27.3 & 84.7 & 65.0 & 47.3 & 31.0 & 24.2 & 20.1 & 17.3 & 116.6 & 105.8 & 88.0 & 67.2 & 53.7 & 44.7 & 38.4 \\
  & \sSthree & \textbf{97.0} & \textbf{86.2} & 67.1 & 49.0 & 38.8 & 32.3 & 28.1 & \textbf{86.7} & \textbf{65.4} & 47.5 & 31.2 & 24.7 & 20.3 & 17.5 & \textbf{119.8} & \textbf{109.2} & \textbf{90.8} & 69.2 & 55.8 & 47.4 & 39.9 \\
  & \sAone & 88.3 & 76.2 & 66.1 & 50.7 & 41.7 & 35.0 & 30.5 & 78.2 & 60.6 & 49.0 & 35.7 & 28.1 & 22.9 & 19.6 & 107.4 & 95.8 & 85.7 & 69.3 & 58.3 & 50.2 & 43.9 \\
  & \sAthree & 87.9 & 78.5 & \textbf{69.8} & \textbf{56.5} & \textbf{47.2} & \textbf{39.8} & \textbf{34.5} & 79.2 & 64.4 & \textbf{53.2} & \textbf{39.3} & \textbf{31.3} & \textbf{25.9} & \textbf{21.9} & 108.0 & 96.5 & 89.6 & \textbf{75.9} & \textbf{65.3} & \textbf{56.4} & \textbf{51.0} \\
\bottomrule
\multicolumn{23}{l}{bold $=$ fastest per column within each group.} \\
\end{tabular}}
\caption{\clrev{\textbf{RTX~3090 edge}, 1 and 2 clients, with the server of Table~\ref{tab:systems_current}. 1 client: tok/s; 2 clients: sum of both clients' tok/s (two edge GPUs sharing one A100).}}
\label{tab:edge3090_current}\label{tab:edge3090}\label{tab:edge3090-2c}  %
\end{table}

\paragraph{\clrev{Two-client shared-server load.}}
\label{app:edge3090_2client}

\edit{\clrev{Every sweep so far serves one edge client per server, whereas a}
central motivation for offloading verification to a shared cloud target is that one
server amortizes its GPU across \emph{many} edge clients. As a first probe of this regime, the RTX~3090 grid was repeated with
two concurrent edge clients sharing one A100 server (two-client group of Table~\ref{tab:edge3090_current}).} \clrev{Dividing the summed throughput by the
matched single-client value gives ${\approx}2.0\times$ for \sSone{} at every RTT and for \sB{}/\sSthree{} from $50$~ms on ($1.6$--$1.7\times$ at $0$~ms)
(the server is mostly idle between their rounds) and $1.60$--$1.99\times$ for \sAthree{}, rising
with the RTT ($1.60$--$1.76\times$ at added $0$~ms, ${\geq}1.9\times$ from $300$~ms) because the
asynchronous schedule already keeps the server busier with one client. The server verifies at batch $1$ throughout, interleaving the two request streams; batching verifications across clients would let one server serve more edges at little cost in aggregate throughput.}

\section{\clrev{Router: Ablations and Extensions}}
\label{app:router}
\clrev{The router results here and in \S\ref{sec:main_results} use per-query latencies from the calibrated profiles of
Eq.~(\ref{eq:router-latency}), applied to each query's prompt and to the output its chosen mode produced, rather than from a separate
end-to-end run of the routed workload.}

\subsection{\edit{Router Behavior}}
\label{sec:router_ablations}\label{sec:composition}

\paragraph{Server usage.} \rrev{An always-cloud system} sends every query to the
cloud, so its server-usage rate is $100\%$ by definition. \rrev{On SPROUT at} the
$500$~ms tail, the router serves only \clrev{$43.3\%$} of queries from the cloud,
\rrev{$2.31\times$ fewer, and those queries carry only $36.1\%$ of the output tokens, so the
verifier processes $2.8\times$ fewer tokens (server time in Appendix~\ref{app:server_cost})}. Realized
accuracy holds at $79.6\%$: the on-device reasoning modes recover the accuracy the cloud
would have supplied, so the offloaded queries lose no accuracy.

\begin{figure}[H]
  \centering
  \includegraphics[width=\linewidth]{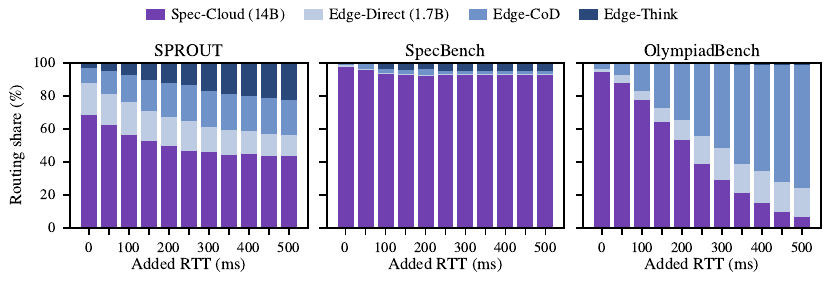}
  \vspace{-0.6em}
  \caption{\clrev{Routing shares vs.\ added RTT on the three benchmarks ($50$~ms steps).}}
  \label{fig:routing_grid}
\end{figure}

\paragraph{\clrev{Per-benchmark routing.}}
\clrev{The split depends on how much of a benchmark the edge can answer well (Figure~\ref{fig:routing_grid}).
On SPROUT the cloud share falls gradually from $68\%$ to $43\%$ as the RTT grows and the two on-device reasoning
modes absorb the difference. On SpecBench the $14$B answers dominate the judge score, so \rrev{$92$--$97\%$} of the queries
stay on the cloud at every RTT and routing changes little. On OlympiadBench the $1.7$B chain-of-draft matches the
non-thinking $14$B target ($54.8$ vs.\ $51.6\%$ correct), so once the round trip costs more than the longer on-device
answer the router moves almost all queries to the edge (cloud share $95\%$ at $0$~ms, $6\%$ at $500$~ms). The router
therefore adapts per query rather than per benchmark: the same rule yields three different regimes.}

\paragraph{Per-group behavior.} The aggregate result conceals
informative per-group structure, shown in Table~\ref{tab:per_group}
at $300$~ms, reported at the \emph{max-accuracy} operating point \rrev{of the router on \sAthree}. At this
point, overall accuracy is $80.5\%$ ($+0.9$~pp over the \rrev{always-$14$B target}), and the
aggregate throughput speedup is \rrev{$1.72\times$ over \sSone}. The router
optimizes a \emph{global} objective, so individual groups
may sit above or below the always-cloud accuracy while the aggregate gains: it
reallocates accuracy from where the edge is competitive to obtain latency savings
where queries are edge-sufficient.

\begin{table}[H]
\centering
\scriptsize
\setlength{\tabcolsep}{3pt}
\renewcommand{\arraystretch}{0.9}
\color{rtrrev}
\begin{tabular}{lrrrc rrrr}
\toprule
& & \multicolumn{2}{c}{\textbf{Acc.\ (\%)}} & & \multicolumn{4}{c}{\textbf{Share (\%)}} \\
\cmidrule(lr){3-4}\cmidrule(lr){6-9}
Group & $n$ & \sSone & \sR & Spd & Cl & Dir & CoD & Th \\
\midrule
GPQA       & 94   & 42.6 & 41.5 & 1.54$\times$ & 74 & 2 & 6 & 17 \\
MuSR       & 123  & 59.3 & 61.8 & 2.09$\times$ & 49 & 2 & 15 & 35 \\
MMLU-Pro   & 1772 & 69.4 & 67.6 & 1.58$\times$ & 78 & 3 & 9 & 10 \\
MATH       & 1462 & 88.1 & 93.2 & 1.57$\times$ & 11 & 3 & 61 & 25 \\
OpenHermes & 2912 & 82.9 & 83.5 & 1.89$\times$ & 57 & 12 & 5 & 27 \\
RAGBench   & 274  & 87.2 & 87.0 & 1.73$\times$ & 70 & 1 & 7 & 22 \\
\midrule
All        & 6637 & 79.6 & 80.5 & 1.72$\times$ & 53 & 7 & 19 & 22 \\
\bottomrule
\end{tabular}
\caption{\rrev{Per-group routing on SPROUT at $300$~ms added latency, max-accuracy operating point (router on \sAthree; overall $80.5\%$, $+0.9$~pp over the always-$14$B target; $1.72\times$ \sSone{} throughput). Acc: \sSone{} (the $14$B target) vs.\ \sR; Spd: speedup over \sSone{} (as in Table~\ref{tab:systems}); Share: routing fractions over \SpecCloud{}, \EdgeDirect{}, \EdgeCoD{}, \EdgeThink{}.}}
\label{tab:per_group}
\end{table}

Three patterns emerge. \emph{(i) Edge-sufficient groups gain speed \emph{and}
accuracy.} OpenHermes---open instruction following, and the largest
group---routes $12\%$ of its queries to \EdgeDirect{} and $27\%$ to
\EdgeThink{}, and outperforms the always-cloud baseline on both axes
($82.9\!\to\!83.5$ at \rrev{$1.89\times$}, rising to \rrev{$2.10\times$} by $500$~ms); MuSR
gains $+2.4$~pp at \rrev{$2.09\times$}. \emph{(ii) On-device reasoning gains accuracy
where the edge outperforms the cloud, without a latency penalty.} On MATH, chain-of-draft is
\emph{more} accurate than the 14B cloud ($93.5\%$ vs.\ $88.1\%$), so the router
sends \rrev{$61\%$} of MATH queries to \EdgeCoD{}, lifting accuracy by $5.1$
points \emph{while} still running \rrev{$1.57\times$} faster at $300$~ms (rising to
\rrev{$2.12\times$} by $500$~ms)---the concise drafts preserve the latency gain while adding accuracy. \emph{(iii) Hard knowledge stays on the cloud.} GPQA and MMLU-Pro remain
cloud-dominant (\rrev{$74\%$, $78\%$}), where the 1.7B cannot match the target; their
residual dips ($-1.1$, \rrev{$-1.8$}) are far outweighed by the MATH, MuSR, and
OpenHermes gains, for an aggregate of $80.5\%$---$+0.9$~pp \emph{above} the
always-cloud baseline.

\subsection{Generalization to Unseen Data}
\label{sec:generalization}

To test whether the routing function transfers beyond its fitting distribution,
we evaluate on OlympiadBench~\cite{he-etal-2024-olympiadbench} ($n\!=\!675$) without
refitting the kNN index. OlympiadBench is both out-of-distribution---no
Olympiad-level problems appear in SPROUT---and substantially harder: the \rrev{always-cloud target}
reaches only $51.6\%$, versus $\sim\!79.6\%$ on SPROUT.
\routeredit{The OlympiadBench columns of Table~\ref{tab:systems} already report the
throughput sweep (up to \clrev{$2.22\times$} at $500$~ms); here we describe how the routing
distribution shifts to achieve it.}
\routeredit{At the fixed matched-accuracy operating point used in the
main-text cost analysis, the measured \SpecCloud{} fractions are
\clrev{$78\%$, $53\%$, and $29\%$ of the queries ($78$/$57$/$31\%$ of the output tokens)} at $100$, $200$, and $300$~ms of added
latency, respectively.}

\routeredit{The router behaves as intended at both ends of the sweep. \emph{At low
RTT} it keeps most queries on the cloud (cloud share \clrev{\rrev{$95\%$} at $0$~ms, $53\%$ at $200$~ms and $29\%$ at
$300$~ms}) and \clrev{stays at parity with running every query on \SpecCloud{} ($0.95$--$0.99\times$ up to \rrev{$200$}~ms)}:
with hard, unfamiliar problems and a cheap network it does not
over-offload to an edge that would lose accuracy. \emph{At high RTT}, as the cloud's
per-query latency climbs (\rrev{to $84$~s on average at $500$~ms}), the router shifts up to \clrev{$74\%$} of the queries to
\EdgeCoD{} and reaches \clrev{$2.22\times$} throughput at iso-accuracy by $500$~ms
(Table~\ref{tab:systems}). This is not a quality sacrifice: router accuracy
holds at $51$--$52\%$ across the sweep\clrev{, and on-device chain-of-draft alone scores $54.8\%$ against the target's $51.6\%$ (Figure~\ref{fig:routing_grid}), because on}
Olympiad-level mathematics, concise on-device reasoning outperforms the cloud
target's direct answer. \rrev{Traded for accuracy instead, the max-accuracy operating point reaches $55.9\%$, $+4.3$~pp above
the target, while still running $1.87\times$ faster than \sSone{} at $500$~ms (Table~\ref{tab:operating_points}).}
The routing function transfers to a harder, unseen
domain with no refitting: it neither over-commits to the edge when the network
is cheap nor fails to exploit it when the network makes the cloud expensive.}

\routeredit{We report SpecBench as an instructive negative result, because it delineates
\emph{when} embedding-based routing helps. On SpecBench the candidate qualities are sharply
asymmetric---the $14$B scores $0.92$ (LLM-judge) while every $1.7$B variant sits at
\rrev{$0.70$--$0.73$}---so the edge rarely substitutes for the cloud. The router recognizes this:
its cloud share stays above $92\%$ across the \emph{entire} RTT sweep (from \rrev{$97.2\%$}
at $0$~ms to $92.7\%$ at $500$~ms), in contrast to SPROUT (\clrev{$68\%\!\to\!43\%$}) and
OlympiadBench (\clrev{$95\%\!\to\!6\%$}). This is graceful degradation, not a failure mode: rather than over-offloading
and losing accuracy, the router \rrev{stays within $0.5$~pp of} $14$B quality by staying on the cloud, so it never
does \emph{worse} than the \SpecCloud{} variant it runs on \rrev{in latency}. }

\subsection{\edit{Router Model Size}}
\label{app:router_size}

\edit{The router's kNN scorer uses a \rrev{Qwen3-Embedding-0.6B} encoder rather than the 1.7B edge draft
model. Two considerations motivate this. First, moving from 0.6B to 1.7B for the routing
decision yields little accuracy change, so the larger model adds no routing quality.
Second---and decisive for end-to-end latency---the 0.6B routing forward can be overlapped
with the 1.7B draft model's prefill, whereas a 1.7B router would serialize behind it and add
end-to-end overhead on every query.}

\subsection{\routeredit{Router Operating Points: Iso- vs.\ Max-Accuracy}}
\label{app:operating_points}

\routeredit{Table~\ref{tab:operating_points} reports the router's two operating points
side by side, both as speedups over \sSone{} (as in Table~\ref{tab:systems}):
\emph{iso-accuracy} (the smallest-latency $\lambda$ whose overall accuracy matches the
always-$14$B target\rrev{; the \sR{} column of Table~\ref{tab:systems}}) and \emph{max-accuracy} (the $\lambda$ that maximizes accuracy). The
accuracy headroom $\Delta$acc---how far the max-accuracy point exceeds the $14$B
target---is strongly dataset-dependent: largest on OlympiadBench ($+4.3$~pp, where the
$14$B target is weakest, ${\approx}51.6\%$), modest on SPROUT (\rrev{$+0.9$--$1.1$}~pp), and absent on
SpecBench (\rrev{$-0.3$ to $-0.5$}~pp, where $14$B is near-ceiling and the $1.7$B models cannot match it,
cf.\ \S\ref{sec:generalization}). On SPROUT and OlympiadBench the
max-accuracy point remains a substantial speedup (\rrev{$1.96\times$ and $1.87\times$} over
\sSone{} at $500$~ms), so exceeding the target's accuracy costs no latency relative to the
cloud.}

\providecolor{rtrrev}{RGB}{200,70,0}\providecommand{\rrev}[1]{\textcolor{rtrrev}{#1}}  %
\begin{table}[H]
\centering\scriptsize
\color{rtrrev}
\setlength{\tabcolsep}{2pt}
\renewcommand{\arraystretch}{0.9}
\caption{\rrev{Router speedup over \sSone{} at the iso-accuracy and max-accuracy operating points (router
on the fastest \SpecCloud{} variant of each cell: \sAthree{}, or \sSthree{} for SpecBench at added~$0$), with the
max-accuracy point's accuracy change over the $14$B target ($\Delta$acc, pp). The iso-accuracy row is the
\sR{} column of Table~\ref{tab:systems}; on SpecBench no operating point reaches the $14$B accuracy, so both rows
are the max-accuracy point.}}
\label{tab:operating_points}\label{tab:operating_points_current}
\begin{tabular}{l l rrrrrr}
\toprule
\textbf{Dataset} & \textbf{Operating point} & \multicolumn{6}{c}{\textbf{Added RTT (ms)}} \\
\cmidrule(lr){3-8}
 & & \textbf{0} & \textbf{100} & \textbf{200} & \textbf{300} & \textbf{400} & \textbf{500} \\
\midrule
\multirow{3}{*}{SPROUT} & Iso-acc ($\times$\sSone) & 1.24 & 1.59 & 1.84 & 1.97 & 2.13 & 2.21 \\
 & Max-acc ($\times$\sSone) & 0.92 & 1.31 & 1.56 & 1.72 & 1.86 & 1.96 \\
 & $\Delta$acc (pp) & $+1.1$ & $+1.0$ & $+0.9$ & $+0.9$ & $+0.9$ & $+0.9$ \\
\midrule
\multirow{3}{*}{SpecBench} & Iso-acc ($\times$\sSone) & 1.12 & 1.33 & 1.50 & 1.50 & 1.47 & 1.47 \\
 & Max-acc ($\times$\sSone) & 1.13 & 1.33 & 1.50 & 1.50 & 1.47 & 1.47 \\
 & $\Delta$acc (pp) & $-0.3$ & $-0.3$ & $-0.4$ & $-0.5$ & $-0.5$ & $-0.5$ \\
\midrule
\multirow{3}{*}{OlympiadBench} & Iso-acc ($\times$\sSone) & 1.13 & 1.30 & 1.37 & 1.65 & 1.98 & 2.22 \\
 & Max-acc ($\times$\sSone) & 0.62 & 0.94 & 1.09 & 1.32 & 1.62 & 1.87 \\
 & $\Delta$acc (pp) & $+4.3$ & $+4.3$ & $+4.2$ & $+4.4$ & $+4.4$ & $+4.3$ \\
\bottomrule
\end{tabular}
\end{table}

\subsection{\clrev{Server Compute and Cost}}
\label{app:server_cost}
\label{app:server_only}  %

\clrev{The asynchronous overlap of drafting with verification costs additional server compute.
Unlike the synchronous baselines, which deepen their proactive tree as the RTT grows and so spend
the same verify budget on more depth (column \rrev{$d_{\max}$} of Table~\ref{tab:systems_current}), the
asynchronous schedule keeps its tree shallow and instead issues \emph{more} verification requests
while a reply is in flight, so it uses $1.8$--$2.3\times$ the verify-forward seconds of the
synchronous schedules at matched RTT (Table~\ref{tab:utilization_current}: OlympiadBench, $100$--$300$~ms added RTT), roughly constant as the
RTT grows while the baselines idle more. Costs follow SpecEdge's cost-efficiency metric---tokens per
dollar $=$ throughput $\div$ hourly rental~\cite{park2025specedge}---with the RTX~4090 edge counted
identically for every schedule. On whole-session rental (billed by wall-clock time) the asynchronous
schedule's faster completion makes the A100 rental per token $16$--$23\%$ cheaper than \sSthree{} at
matched RTT; under a forward-only lower bound (a fully batched, multi-client server) its extra
verifier work makes it ${\sim}50\%$ more expensive per token. These cloud-side costs apply only to the queries routed to the cloud:
at the matched-accuracy operating point the router sends $78$/$57$/$31\%$ of the
OlympiadBench output tokens to the cloud at $100$/$200$/$300$~ms, and
scaling the \sAthree{} values by that share gives $149$/$111$/$61$~s of verifier compute and
\$$17$/$16$/$10$ per $1$M tokens under rental---below every always-cloud baseline---and the lowest
forward-only cost from $200$~ms on.}

\begin{figure}[H]
  \centering
  \includegraphics[width=\linewidth]{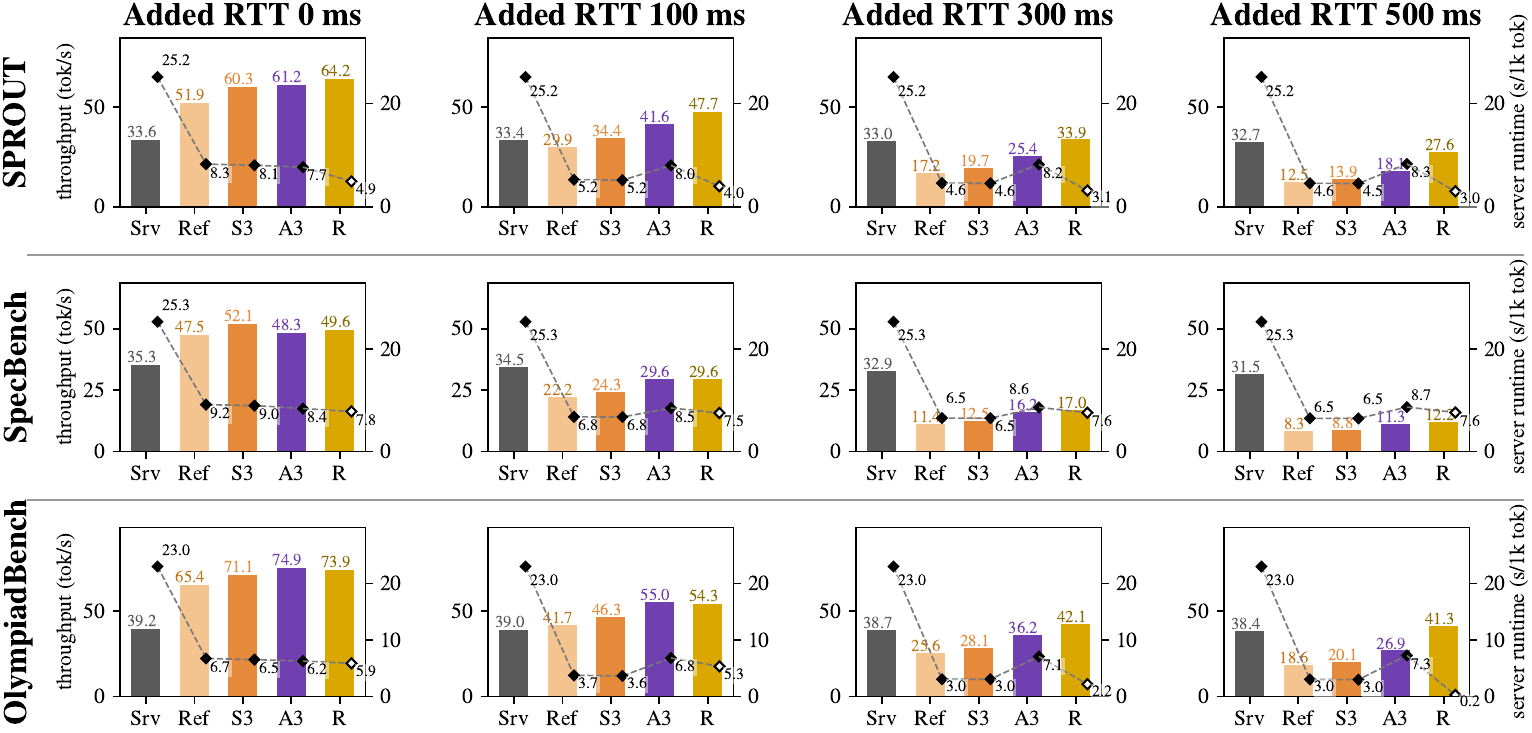}
  \vspace{-0.8em}
  \caption{\clrev{Decoding throughput (bars) and server runtime per $1$k output tokens (\(\blacklozenge\)). Srv $=$ server-only speculative decoding, whose draft forwards also run on the server; \sR's server time (\(\lozenge\)) $=$ \sAthree's per-token server time $\times$ the router's cloud token share.}}
  \label{fig:server_cost_all}
\end{figure}

\clrev{\textbf{Server-only speculative decoding.} At long RTT one might instead keep both draft and target on the cloud and skip the
round trip. Figure~\ref{fig:server_cost_all} includes this baseline (Srv: draft and target on one A100 in one process, the draft tree verified
right after it is built; same models, budget and greedy decoding; its draft forwards, $14.8$~ms per $32$-beam step, count as server time). It reaches $33.6$--$39.2$~tok/s
with the \sSone{} draft stack ($40.1$/$39.9$/$44.1$ \rrev{on SPROUT/SpecBench/OlympiadBench with the \sSthree{} one}, which is $0.59$--$0.76\times$ the best edge--cloud schedule at zero added
latency), and since it pays no round trip it overtakes \sAthree{} in raw throughput from about $50$~ms (SpecBench)\rrev{, $150$~ms (SPROUT)} \rrev{and} $200$~ms (OlympiadBench)
of added latency. The right axis of the figure shows the cost: the single A100 is busy for the whole run, so per $1$k output tokens it spends
about $3\times$ the server time of the synchronous schedules at zero added latency and \rrev{about} $4$--$8\times$ from $100$~ms on (their verify work shrinks as the round trip grows), and $2.9$--$3.7\times$ that of \sAthree. Offloading the draft to the edge is what
makes the server verify-only, which is the SpecEdge premise; our design trades a round trip---hidden by the asynchronous overlap or
avoided outright by the router---for that server capacity.}
\begin{table}[H]
\centering\footnotesize
\setlength{\tabcolsep}{2.5pt}
\renewcommand{\arraystretch}{0.9}
\color{clrevblue}
\begin{tabular}{lll cccccc}
\toprule
\textbf{Metric} & \textbf{Added} & \textbf{Eff.\ RTT} & \sSone & \sB & \sSthree & \sAone & \sAthree & \sR \\
\midrule
Server compute (s) & 100 & 115 & 107 & 105 & 103 & 201 & 191 & 149 \\
 & 200 & 218 & 92 & 90 & 90 & 207 & 195 & 111 \\
 & 300 & 321 & 88 & 88 & 89 & 212 & 198 & 61 \\
\addlinespace
Rental (\$/1M tok) & 100 & 115 & 29 & 27 & 26 & 26 & 22 & \textbf{17} \\
 & 200 & 218 & 39 & 36 & 35 & 33 & 28 & \textbf{16} \\
 & 300 & 321 & 48 & 45 & 44 & 41 & 34 & \textbf{10} \\
\addlinespace
Compute (\$/1M tok) & 100 & 115 & 6.5 & 6.3 & \textbf{6.2} & 10.0 & 9.4 & 7.4 \\
 & 200 & 218 & 6.7 & 6.4 & 6.3 & 10.8 & 10.0 & \textbf{5.7} \\
 & 300 & 321 & 7.2 & 7.0 & 6.9 & 11.5 & 10.6 & \textbf{3.2} \\
\bottomrule
\end{tabular}
\caption{\clrev{Server compute and cost on OlympiadBench (RTX 4090 edge, one client): verifier-forward time per $32$-query run, session rental cost per $1$M tokens (A100 \$4.05/h $+$ RTX 4090 \$0.35/h by wall clock) and the forward-only lower bound; \sR{} $=$ \sAthree{} scaled by the router's cloud token share ($78$/$57$/$31\%$ at $100$/$200$/$300$~ms); bold $=$ cheapest.}}
\label{tab:utilization_current}\label{tab:utilization}  %
\end{table}

\section{\clrev{Extended Related Work}}
\label{app:related}
\clrev{Same-request runahead---drafting past a block that is still being verified---originates in
PipeInfer~\cite{10.1109/SC41406.2024.00046}, which keeps speculating while earlier micro-batches
traverse a pipeline-parallel target and cancels invalidated work. AMUSD~\cite{McDanel2024AMUSDAM}
and PEARL~\cite{liu2025pearl} overlap drafting and verification on a single node,
SpecPipe~\cite{yin2025specpipe} streams a fixed-width tree level by level into a pipeline-parallel
target, and PipeSD~\cite{han2026pipesd} pipelines edge drafting with cloud verification. In all of
these the runahead depth is fixed by the hardware (a round or two) and rejected work is cancelled
or flushed. SwiftSpec~\cite{10.1145/3779212.3790246} scales the idea over NVLink with a disaggregated draft model \rrev{whose draft side keeps growing the tree while a batch is verified and then re-roots it at the verified
token}; none treats the WAN RTT as the variable \rrev{or keeps several hedged trees in flight across a round trip}.}

\clrev{Beyond SpecEdge, edge--cloud speculative decoding has been studied along two other axes.
One shares or splits the verifier: DSSD~\cite{ning2025dssd} splits verification between device and
edge, SLED~\cite{10.1145/3769102.3770608} and WISP~\cite{10.1145/3805653} serve many edge drafters from a shared
verifier, and CoSine~\cite{11571388} decouples drafting and verification across collaborating
draft nodes. The other reduces the bytes per round: conformal
sparsification~\cite{bhattacharjee2025conformal}, quantize-sample-and-verify~\cite{zhang2026quantize},
and \rrev{cooperative cache sharing with incremental synchronization}~\cite{11374072}. All of these keep a synchronous
draft--verify schedule per request. AsymSpec~\cite{yang2026asymspec} goes further and
\emph{forbids} same-request runahead---at most one open block per request---filling verification
waits with other requests to raise multi-tenant throughput under asymmetric bandwidth, and
Lyu et al.~\cite{lyu2026distance} show analytically and empirically that edge--cloud speculative
decoding reduces latency only inside a bounded RTT window, outside of which co-located speculative
decoding wins. \name{} targets that regime for a single request: it keeps $2$--$9$ fixed-budget
trees in flight as the added RTT grows from $0$ to $500$~ms, contains the wasted work with a
committed-prefix gate, continuation hedging, and lineage-preserving rebuild, and its router falls
back to on-device modes once the cloud path stops paying off.}

\end{document}